# The Nonverbal Syntax Framework: An Evidence-Based Tiered System for Inferring Learner States from Observable Behavioral Cues

Sherzod Turaev[1*], Mary John[2], Jaloliddin Rustamov[1], Zahiriddin Rustamov[1], Saja Aldabet[1], Nazar Zaki[1], Khaled Shuaib[3]

[1] Department of Computer Science & Software Engineering, College of Information Technology, United Arab Emirates University, Al Ain, Abu Dhabi, United Arab Emirates

[2] Academic Support Department, Abu Dhabi Polytechnic, Abu Dhabi, United Arab Emirates

[3] Department of Information Systems and Security, College of Information Technology, College of Information Technology, United Arab Emirates University, Al Ain, Abu Dhabi, United Arab Emirates

[*] Corresponding Author: Sherzod Turaev (sherzod@uaeu.ac.ae)

## Abstract

Understanding learners' internal states, both cognitive and affective, is essential for adaptive educational systems and effective teaching. Although much research has examined relationships between observable nonverbal cues and internal states, there is no comprehensive framework to systematically organize these insights with proper evidence calibration. This paper presents the Nonverbal Syntax Framework, an evidence-based system for mapping behavioral cues to learner states, built on a systematic literature review of 908 empirical studies that yielded 17,043 cue-state mappings (Turaev et al., 2026).

The framework addresses three critical challenges in applying research findings to practice: (1) *terminological fragmentation*, where identical behaviors are described inconsistently across studies; (2) *evidence heterogeneity*, where relationships range from single observations to extensively replicated findings; and (3) *state ambiguity*, where similar behavioral patterns can indicate multiple internal states. Through systematic normalization procedures, we consolidated 5,537 raw state labels into 2,010 canonical states (63.7% reduction) and 11,521 raw cue descriptions into 6,434 normalized cues (44.2% reduction), organized across nine behavioral channels.

Central to our approach is a dual-evidence assessment system that separately evaluates *Component Evidence* (how well-studied individual cues and states are across all associations) and *Relationship Evidence* (how many independent studies document each specific cue-state link). This distinction is critical: we found that 52% of relationships rated "Very High" for confidence based on well-studied components are documented in only a single paper. By separating these metrics, the framework enables appropriately standardized inference rather than overconfident application of preliminary findings.

The framework comprises four interconnected levels: (1) a *Cue Vocabulary* cataloging 6,434 behavioral indicators with observable/instrumental classification; (2) *State Clusters* associating each of 2,010 states with its indicative cues; (3) *State Profiles* providing multimodal behavioral signatures with actionable cue specifications; and (4) *Discriminative Analysis* identifying cues that distinguish between 1,215 potentially confusable state pairs.

The framework identifies 480 actionable relationships with R1–R4 evidence (three or more independent papers) representing the consolidated, replicated core of six decades of research, covering 35.5% of all documented mappings across 47 key learning states and 111 distinct behavioral indicators. Of the remaining unique relationships, 91.5% (9,653 single-paper findings) represent exploratory hypotheses for future replication. The Nonverbal Syntax Framework provides researchers with an empirically grounded foundation for identifying knowledge gaps, practitioners with evidence-based tools for state inference, and technologists with validated features for multimodal detection systems.

# 1. INTRODUCTION

Effective teaching and learning depend critically on the ability to recognize and respond to learners' internal states. When students experience confusion, timely intervention can transform a moment of struggle into productive learning (D'Mello & Graesser, 2012). Conversely, undetected boredom or frustration can lead to disengagement and learning failure (Baker et al., 2010). As educational technologies become increasingly sophisticated, the demand for systems capable of automatically detecting and responding to learner states has grown substantially (Calvo & D'Mello, 2010).

Learner internal states encompass both cognitive dimensions, such as attention, engagement, and cognitive load, and affective dimensions, including confusion, frustration, boredom, and delight (Pekrun, 2006). These states are not merely incidental to learning; they play a fundamental role in shaping learning outcomes. Research has demonstrated that affective states influence memory consolidation (Tyng et al., 2017), problem-solving strategies (Isen, 2001), and persistence in the face of difficulty (Dweck, 2006). Cognitive states such as attention and engagement directly determine which information is processed and retained (Fredricks et al., 2004).

We naturally infer others' internal states through their observable behavioral cues. Facial expressions convey emotional valence and arousal (Ekman, 1992); eye movements reveal attention allocation and cognitive processing (Just & Carpenter, 1980); body posture signals engagement or withdrawal (Harrigan, 2005); and vocal characteristics indicate stress, confidence, or uncertainty (Scherer, 2003). This natural capacity for "reading" nonverbal behavior has inspired decades of research aimed at computational detection of affective and cognitive states (Picard, 1997).

Advances in affordable sensors and computer vision technologies have made it increasingly feasible to capture multimodal behavioral data in educational settings (Woolf et al., 2009). Recent progress in deep learning has significantly improved the accuracy of automatic affect recognition, achieving near-human performance on benchmark datasets (Li & Deng, 2022). This technological capability has driven extensive research on mapping observable cues to underlying states, particularly in online and remote learning contexts, which expanded significantly following the COVID-19 pandemic (Dewan et al., 2019; Sharma et al., 2020). The post-pandemic shift toward hybrid instruction and the rapid integration of generative AI tools into learning environments have further reshaped the contexts in which learner states must be observed, raising the question of whether cue-state relationships established in earlier classroom-based studies remain valid in contemporary digital settings. Two considerations mitigate this concern. First, the channels underlying nonverbal communication, including facial action, gaze, posture, and prosody, are anchored in human physiology and emotional expression and are therefore unlikely to have changed materially over the period covered by this synthesis. Second, the empirical evidence base itself has shifted markedly toward recent research, with 56.8% of the 908 studies underlying the present framework published between 2020 and 2025, ensuring that contemporary learning contexts are well represented. At the same time, certain context-bound cues, such as screen-oriented postures or interaction-log behaviors characteristic of online learning, have emerged or gained prominence only in recent years, while traditional classroom indicators such as hand-raising may be diminishing in some settings; these temporal shifts are revisited as a limitation in Section 4.

Despite decades of research, the field lacks a unified framework for understanding cue-state relationships. Three critical problems impede the translation of research findings into practical application:

*First, terminological fragmentation.* The same behavioral phenomenon may be described differently across studies: “forward lean,” “leaning toward the screen,” and “body orientation toward the task” may all refer to similar postural configurations. Similarly, internal states suffer from inconsistent labeling: “engaged concentration,” “flow,” “deep engagement,” and “absorbed attention” often describe overlapping constructs. This inconsistency makes it difficult to synthesize findings across studies or build cumulative knowledge (Bosch et al., 2015).

*Second, evidence heterogeneity.* Relationships between cues and states vary dramatically in their empirical support. Some associations, such as the link between frowning and confusion, have been documented across numerous independent studies employing diverse methodologies (Grafsgaard et al., 2013). Others rely on single observations or small-sample studies. Without a systematic method for assessing evidence strength, practitioners cannot calibrate their confidence across different cue-state mappings.

*Third, state ambiguity.* Many nonverbal cues are associated with multiple, sometimes contradictory, internal states. Furrowed brows may indicate confusion, concentration, or frustration (Craig et al., 2004). Reduced eye contact might signal boredom, cognitive load, or social anxiety. This many-to-many mapping between cues and states poses substantial challenges for both automatic detection systems and human observers.

Previous efforts to address these challenges have taken several forms. Emotion recognition systems typically focus on the six basic emotions derived from Ekman’s work (Ekman & Friesen, 1971), but these categories poorly capture learning-relevant states like confusion or engagement (D’Mello & Graesser, 2012). Engagement detection systems often define states narrowly for specific contexts, limiting generalizability (Whitehill et al., 2014). Deep learning approaches have achieved impressive accuracy on benchmark datasets but often function as “black boxes” that provide little insight into which behavioral features drive predictions (Zeng et al., 2009), and frequently fail to generalize across contexts, populations, or recording conditions (Kołakowska et al., 2017).

Importantly, existing approaches typically treat cue-state relationships as binary (present/absent) or probabilistic without distinguishing between well-established relationships and preliminary findings. This conflation of strong and weak evidence can mislead system designers and researchers. Recent reviews have called for more systematic approaches to synthesizing the growing body of research on behavioral indicators of learner states (Sharma & Hannafin, 2007; Arguel et al., 2019).

This paper introduces the Nonverbal Syntax Framework, an evidence-based system designed to address fragmentation and enable the principled translation of research findings into practice. The framework is built upon a comprehensive systematic literature review that synthesized findings from 908 empirical studies, yielding 17,043 cue-state mappings spanning six decades of research (1966–2025); full methodological details of the review are provided in Turaev et al. (2026).

The term *Nonverbal Syntax* reflects our conceptualization of behavioral cues as elements of a grammar: just as words combine according to syntactic rules to convey meaning, nonverbal cues combine in patterned ways to signal internal states. A single cue, like a single word, carries limited information and admits multiple interpretations. However, when multiple cues co-occur across behavioral channels, facial expression combined with body posture, gaze pattern, and vocal characteristics, they form "behavioral sentences" that constrain interpretation and enable more reliable state inference. Understanding this "syntax" enables practitioners and automated systems alike to move beyond single-cue detection toward pattern-based recognition.

The proposed framework provides:

1. *Terminological normalization* through systematic consolidation of state labels (from 5,537 to 2,010 canonical states; 63.7% reduction) and cue descriptions (from 11,521 to 6,434 normalized cues; 44.2% reduction).
2. *Dual-evidence assessment* that separately evaluates how well-studied individual cues and states are (Component Evidence Tiers) and how well-documented specific cue-state links are (Relationship Evidence Tiers).
3. *Hierarchical organization* across four interconnected levels: Cue Vocabulary, State Clusters, State Profiles, and Discriminative Analysis.
4. *Practical inference support* through actionability classification and diagnostic tools that identify the 480 well-replicated relationships suitable for confident application.

This paper makes five primary contributions:

1. *A comprehensive cue vocabulary* cataloging 6,434 normalized behavioral indicators organized across nine channels (Facial, Eye, Head, Body, Gesture, Voice, Physiological, Behavioral, and Multimodal), with evidence tiers indicating replication strength and classifications distinguishing observable from instrumental cues.
2. *State profiles with multimodal signatures* describing how each of 2,010 internal states manifests across behavioral channels, including detailed behavioral specifications for key learning states (confusion, frustration, boredom, engagement, attention).
3. *A dual-evidence assessment system* that separately evaluates Component Evidence (how well-studied individual cues and states are) and Relationship Evidence (how many independent papers document specific cue-state links). This distinction reveals that 52% of "Very High" confidence relationships based on well-studied components are actually documented by only a single paper, a critical finding for appropriate inference calibration.

4. *Actionability classification* distinguishing between generic cues requiring further specification (e.g., “facial expressions”) and specific, observable behaviors that practitioners can directly detect (e.g., “scratching head,” “sighing”). The 480 actionable relationships with R1–R4 evidence (three or more papers) cover 35.5% of all documented mappings, representing the consolidated core of six decades of research.
5. *Discriminative analysis* identifying 1,215 potentially confusable state pairs and the specific cues that distinguish them, addressing the state ambiguity problem with explicit evidence tiers for each discriminative cue.

Combined, these contributions provide not only a research resource but also practical tools, including diagnostic reference materials, detailed behavioral signatures, and structured lookup procedures that educators and system designers can apply directly in observation and detection contexts.

The remainder of this paper is organized as follows: Section 2 describes the proposed methodology, including a summary of the systematic literature review data foundation, normalization procedures for states and cues, and the dual-evidence assessment system unique to this framework. Section 3 presents the Nonverbal Syntax Framework across four levels, including detailed behavioral signatures for key learning states and worked examples demonstrating practical inference. Section 4 discusses the research and practical implications, compares this framework with existing methods, and outlines limitations. Finally, Section 5 summarizes the main points for researchers and practitioners.

## 2. METHODS

This section describes the methodology underlying the Nonverbal Syntax Framework. The framework is based on a systematic literature review of 908 papers, yielding 17,043 cue-state mappings; complete details of the search strategy, eligibility criteria, screening process, and data extraction procedures are provided in Turaev et al. (2026). Here, we summarize the empirical foundation and present the methodological contributions unique to this framework: normalization procedures that address terminological fragmentation; a dual-evidence assessment system that enables appropriately standardized inference; an actionability classification that bridges research and practice; and the four-level framework architecture.

Table 1 summarizes the corpus underlying the Nonverbal Syntax Framework. From an initial pool of 1,987 records identified through the database searches, 921 studies met the inclusion criteria, of which 908 yielded extractable cue-state mappings totaling 17,043 unique relationships. The corpus spans nearly six decades of research (1966–2025), with the majority of studies (56.8%) published in the most recent five-year window (2020–2025), reflecting both the rapid expansion of affective computing in educational contexts and the surge of interest in remote and hybrid learning environments following the COVID-19 pandemic.

### 2.1 Empirical Foundation: Systematic Literature Review

We conducted a systematic literature review following PRISMA guidelines (Page et al., 2021). The review searched six electronic databases (Semantic Scholar, Google Scholar, ACM Digital Library, PubMed, ScienceDirect, and IEEE Xplore) using a dual search strategy combining semantic and keyword-based approaches. No publication date restrictions were applied, resulting in coverage from 1966 to 2025. Full methodological details, including search terms, eligibility criteria, and the multi-stage validation protocol using dual-model extraction, are reported in Turaev et al. (2026).

**Table 1.** *Systematic Review Corpus Summary*

| Metric | Value |
|---|---|
| Records identified | 1,987 |
| Studies meeting inclusion criteria | 921 |
| Studies with extractable mappings | 908 |
| Total cue-state mappings | 17,043 |
| Publication coverage | 1966–2025 |
| Papers published 2020–2025 | 56.8% |

Figure 1 presents the PRISMA flow diagram; the complete screening process is detailed in Turaev et al. (2026).

## 2.2 Normalization Procedures

Synthesizing research across this literature required addressing considerable terminological variability. The same behavior may be described as “furrowed brow,” “brow lowering,” “corrugator activation,” or “AU4,” depending on disciplinary conventions and measurement approaches. Similarly, internal states exhibit labeling inconsistency: “engaged,” “engagement,” “engaged attention,” “behavioral engagement,” “cognitive engagement,” and “task engagement” may refer to related or identical constructs. Without normalization, these variants would be treated as distinct phenomena, fragmenting the evidence base and obscuring patterns of replication.

We developed hierarchical normalization dictionaries for both states and cues through three components:

1. *Synonym mapping:* Semantically equivalent terms consolidated under standardized labels (e.g., “forward lean,” “leaning forward,” “postural orientation toward” → “forward lean”).
2. *Action Unit decoding:* Facial Action Coding System codes translated to observable behavioral descriptions based on Ekman and Friesen’s (1978) FACS manual (e.g., “AU1+2” → “inner and outer brow raise”).
3. *Specificity classification:* Distinction between specific, well-defined behaviors (e.g., “lip corner pull,” “gaze aversion”) and general behavioral categories (e.g., “positive facial expression,” “body movement”).

The most extensively normalized state was “engagement,” with 666 raw variants consolidated into a single canonical category. Cue normalization was more conservative to preserve behavioral specificity. The complete normalization dictionaries are provided as Supplementary Materials. Table 2 summarizes the normalization statistics.

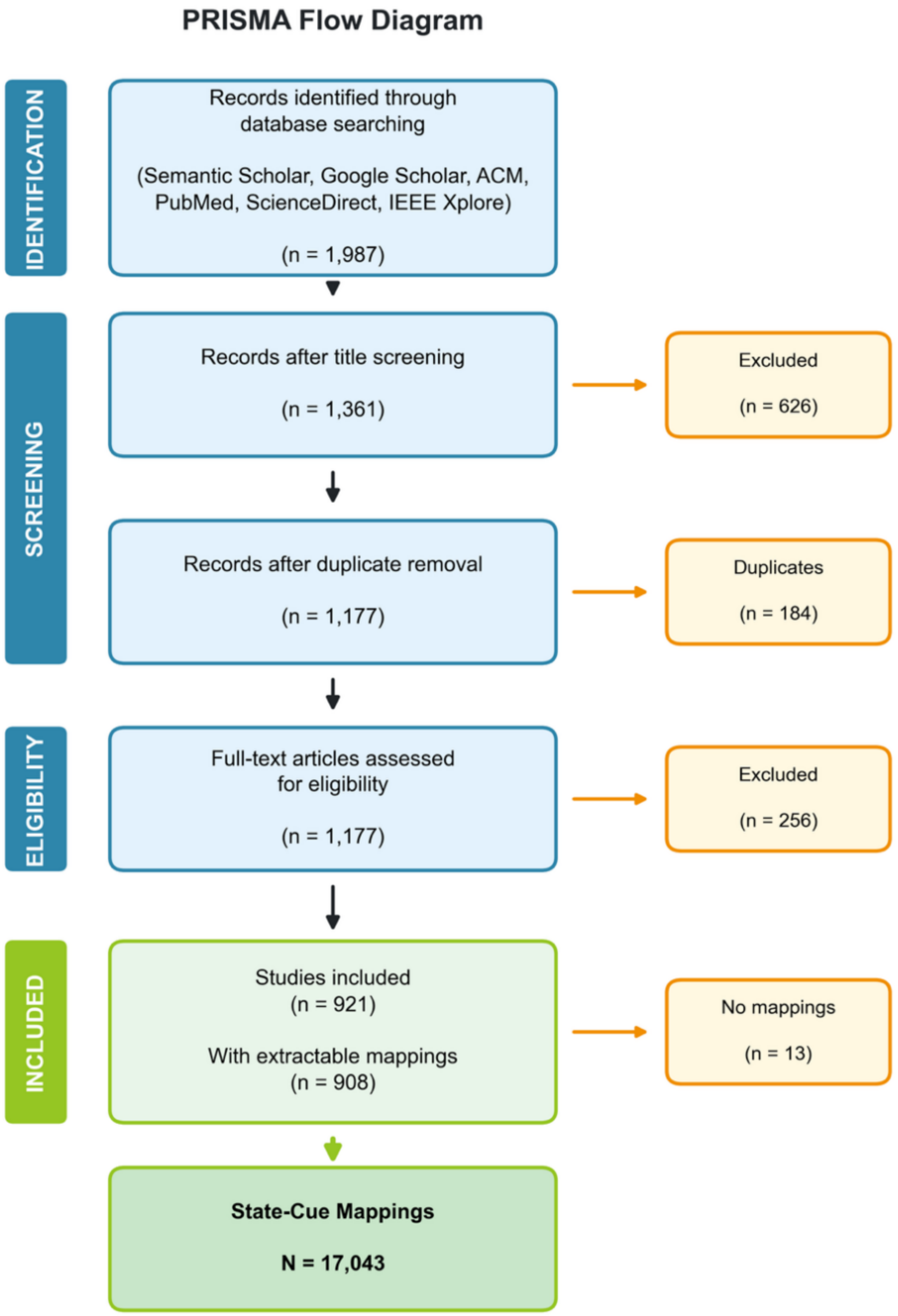

***Figure 1.*** *PRISMA flow diagram of the systematic review. From an initial pool of 1,987 records identified through database searches, 626 were excluded after title screening and 184 were removed as duplicates, yielding 1,177 records for full-text eligibility assessment. After excluding 256 records during eligibility review, 921 studies met the inclusion criteria; of these, 908 yielded extractable cue-state mappings totaling 17,043 relationships, while 13 contained no extractable mappings. The figure reports the mapping count as N = 17,043 based on the screening-phase audit. During normalization, four text-based mappings (textual interactions measured via BERT embeddings) from one study were excluded as falling outside the nonverbal scope, yielding 17,039 rows in the normalized supplementary dataset; all analyses in this paper use the pre-normalization count of 17,043 unless otherwise noted.*

**Table 2.** *Normalization Summary*

| Dimension | Raw Labels | Normalized | Reduction |
|---|---|---|---|
| States | 5,537 | 2,010 | 63.7% |
| Cues | 11,521 | 6,434 | 44.2% |

An important distinction emerged during normalization: some cues are directly observable by human observers without instrumentation, while others require instrumental measurement. Table 3 summarizes this classification.

**Table 3.** *Cue Observability Classification*

| Category | Description | Example Channels | Example Cues |
|---|---|---|---|
| Directly Observable | Detectable by trained human observers | Facial, Head, Body, Gesture, Voice | smile, head nod, slouching, sighing |
| Instrumentally Measured | Requiring sensors or logging | Physiology, Behavioral | skin conductance, heart rate, keystroke patterns |
| Mixed | Some variants observable, others instrumental | Eye Movements | gaze direction (observable) vs. saccade velocity (instrumental) |

This distinction enables practitioners to filter the framework to cues appropriate for their detection context.

## 2.3 Dual-Evidence Assessment System

A critical methodological contribution is the dual-evidence assessment system, which addresses a fundamental limitation in existing evidence synthesis: the conflation of component-level and relationship-level evidence. The remainder of this subsection unpacks this idea in five steps. We first articulate the core problem that motivates the dual-evidence approach, then introduce the two complementary tiers (Component Evidence and Relationship Evidence) along with the rules used to combine them, and finally examine the empirical divergence between the two metrics and its implications for how actionable relationships are interpreted in the framework.

***The Problem: Component vs. Relationship Evidence.*** Traditional confidence ratings assess how well-studied the individual components (cues and states) are across the literature. However, the assumption that a high component-level rating necessarily implies a well-replicated relationship is not always warranted: a cue and a state that have each been studied many times in isolation may have been linked together in only a single paper. Aggregating component-level evidence into a single confidence score therefore risks overstating the replication strength of any specific cue-state combination, particularly when the goal is to identify relationships robust enough to inform practice or to seed automated detection systems.

***Component Evidence Tiers.*** Component Evidence Tiers indicate the overall research attention received by individual cues and states across all their associations. The tier thresholds were chosen to reflect commonly used cut-offs in evidence synthesis: a single observation marks the exploratory frontier (Tier 5), two studies indicate emerging support that requires further validation (Tier 4), three to four studies provide reasonable empirical backing (Tier 3), and five or more studies move a construct into the moderately or extensively replicated range (Tiers 2 and 1). Because states and cues differ in their typical paper counts, the cut-offs for Tier 1 are calibrated separately for the two dimensions: ≥20 papers for states and ≥10 papers for cues. Tables 4 and 5 present the resulting tier distributions, which reveal a heavily skewed evidence base in which a small number of core constructs account for most of the research attention while the long tail of states and cues has been examined in only a single study.

**Table 4.** *State Component Evidence Distribution*

| Tier | Criterion | Count | % | Interpretation |
|---|---|---|---|---|
| Tier 1: Strong | ≥20 papers | 34 | 1.7% | Core construct, extensively studied |
| Tier 2: Moderate | 10–19 papers | 8 | 0.4% | Reliable, solid empirical support |
| Tier 3: Supported | 5–9 papers | 21 | 1.0% | Reasonable empirical backing |
| Tier 4: Emerging | 2–4 papers | 118 | 5.9% | Preliminary, requires validation |
| Tier 5: Exploratory | 1 paper | 1,829 | 91.0% | Single-study observation |

**Table 5.** *Cue Component Evidence Distribution*

| Tier | Criterion | Count | % | Interpretation |
|---|---|---|---|---|
| Tier 1: Strong | ≥10 papers | 64 | 1.0% | Core indicator, extensively studied |
| Tier 2: Moderate | 5–9 papers | 45 | 0.7% | Reliable, solid empirical support |
| Tier 3: Supported | 3–4 papers | 92 | 1.4% | Reasonable empirical backing |
| Tier 4: Emerging | 2 papers | 222 | 3.5% | Preliminary, requires validation |
| Tier 5: Exploratory | 1 paper | 6,011 | 93.4% | Single-study observation |

***Combined Confidence.*** Component evidence on its own captures only how thoroughly each side of a cue-state pair has been studied; it does not capture the strength of the link between the two. To produce a single confidence label for each relationship that nevertheless rests on both component tiers, we adopt a categorical combination rule rather than an averaged numerical score. The categorical approach has two advantages: first, it preserves the asymmetry between strong and weak components, so that pairing a Tier 1 cue with a Tier 5 state never produces a misleadingly moderate score; second, it makes the combination transparent and reproducible, since every relationship can be classified by inspecting the two tier labels rather than by computing a derived numeric value. Table 6 presents the resulting combination rules, which range from “Very High” (when both components fall in Tiers 1 or 2) to “Very Low” (when both fall in Tiers 4 or 5).

**Table 6.** *Combined Confidence Calculation*

| Combined Confidence | Rule |
|---|---|
| Very High | Both cue and state at Tier 1 or Tier 2 |
| High | One component at Tier 1–2, other at Tier 3 |
| Moderate | Both at Tier 3, or one at Tier 1–2 with other at Tier 4 |
| Low | One at Tier 4–5, other at Tier 3 or better |
| Very Low | Both components at Tier 4 or Tier 5 |

***Relationship Evidence Tiers.*** Relationship Evidence Tiers directly assess how many independent papers document each specific cue-state link, which is the critical metric for replication. Table 7 presents the distribution. The same six-level structure used for component tiers is applied here, ranging from R1 (≥20 papers, extensively replicated) to R6 (single-paper, exploratory). Crucially, a relationship at R1–R4 can be interpreted as having genuine cross-study support for the specific cue-state link itself, independent of how often the constituent cue and state have appeared in other associations.

**Table 7.** *Relationship Evidence Distribution*

| Tier | Criterion | Count | % | Interpretation |
|---|---|---|---|---|
| R1: Strong | ≥20 papers | 47 | 0.4% | Extensively replicated |
| R2: Substantial | 10–19 papers | 63 | 0.6% | Well-documented |
| R3: Moderate | 5–9 papers | 137 | 1.3% | Solid replication |
| R4: Supported | 3–4 papers | 233 | 2.2% | Reasonable evidence |
| R5: Emerging | 2 papers | 420 | 4.0% | Preliminary |
| R6: Exploratory | 1 paper | 9,653 | 91.5% | Single-study hypothesis |

Actionable relationships (R1–R4) total 480 unique cue-state pairs, representing 4.5% of the unique relationships in the corpus.

***Critical Finding: Evidence Divergence.*** Cross-tabulating Combined Confidence with Relationship Evidence shows a significant difference between the two metrics, summarized in Table 8.

Over half (52%) of “Very High” combined confidence relationships are documented by only a single paper. This finding underscores why dual-evidence assessment is essential: a relationship can sit at the top of the component-confidence hierarchy and still rest on a single empirical observation of the specific cue-state link itself.

**Table 8.** *Combined Confidence vs. Single-Paper Relationships*

| Combined Confidence | Total | Single Paper (R6) | % Single Paper |
|---|---|---|---|
| Very High | 1,426 | 746 | 52.3% |
| High | 831 | 541 | 65.1% |
| Moderate | 6,362 | 6,120 | 96.2% |
| Low | 449 | 425 | 94.6% |
| Very Low | 1,485 | 1,485 | 100.0% |

***Contextualizing Actionable Relationships.*** While 480 actionable relationships represent only 4.5% of unique cue-state pairs, this reflects the expected long-tailed distribution in scientific evidence, in which a small core of extensively replicated findings carries the majority of evidential weight while a long tail of single-study observations forms hypotheses for future investigation. The choice of three papers as the lower bound for actionability is corroborated post-hoc by formal distributional analysis: a maximum-likelihood power-law fit to the relationship paper-count distribution returns a data-driven cutoff $x_{\min} = 3$ that coincides exactly with this methodological threshold (see Section 4.1.3 and Figure 7). Table 9 summarizes the coverage of this actionable core.

**Table 9.** *Actionable Relationship Coverage*

| Metric | Value |
|---|---|
| Actionable relationships (R1–R4) | 480 (4.5% of unique pairs) |
| Total mappings covered | 6,049 (35.5% of all evidence) |
| States represented | 47 (including all key learning states) |
| Cues represented | 111 distinct behavioral indicators |
| Channels covered | 8 of 9 |

The 480 actionable relationships represent the core, repeatedly validated findings from six decades of research, examined across 908 papers. The 9,653 single-study relationships are exploratory findings that generate hypotheses for future replication.

## 2.4 Actionability Classification

To bridge the gap between research findings and practical application, we classified cues by their *actionability*, the degree to which they specify observable behaviors that practitioners can directly detect and record.

Actionability is conceptually distinct from evidence: a cue may be extensively documented in the literature yet too generic to be directly observable, while a less-studied cue may be entirely actionable in classroom or screen-based contexts. Generic categories such as “facial expressions” or “body movement” are reported frequently in the literature but provide little guidance on what an observer should actually look for. To make this distinction explicit, we classified each cue along a four-level actionability spectrum that reflects the specificity of the behavioral description and the feasibility of direct observation. Highly actionable cues describe a specific, immediately observable behavior such as “scratching head” or “yawning”; moderately actionable cues describe a reasonably specific behavior that an observer can recognize given brief guidelines, such as “leaning forward” or “fidgeting”; weakly actionable cues describe broad behavioral categories that require further specification before they can be operationalized; and non-actionable cues are either generic descriptors such as “facial expressions” or instrumental signals such as “EEG” that cannot be observed without specialized equipment. Table 10 summarizes the four levels with definitions, examples, and their intended practical use.

**Table 10.** *Actionability Classification*

| **Level** | **Description** | **Example** | **Practical Use** |
|---|---|---|---|
| Highly Actionable | Specific, directly observable | “scratching head,” “yawning” | Direct observation |
| Moderately Actionable | Reasonably specific | “leaning forward,” “fidgeting” | Observation with guidelines |
| Weakly Actionable | Broad category | “body movement” | Requires specification |
| Non-Actionable | Generic or instrumental | “facial expressions,” “EEG” | Cannot observe directly |

Because many literature-reported cues are generic categories (e.g., “facial expressions”) rather than specific observable behaviors, the actionability classification was applied not only to the 6,371 original mappings that already named specific cues but also to 18,338 specific behaviors inferred from generic categories through LLM-assisted extraction (e.g., “furrowed brow” and “lip compression” inferred from “facial expressions”), yielding an expanded set of N = 24,709 cue-level assessments. This expanded set provides a more complete picture of actionability across the full cue vocabulary.

Table 11 presents the distribution of mappings across actionability levels.

**Table 11.** *Actionability Distribution of Mappings*

| **Level** | **Count** | **%** |
|---|---|---|
| Highly Actionable | 12,882 | 52.1% |
| Moderately Actionable | 3,908 | 15.8% |
| Weakly Actionable | 253 | 1.0% |
| Non-Actionable | 7,666 | 31.0% |

Over two-thirds of mappings (67.9%) involve cues that are highly or moderately actionable, enabling their application in observation-based contexts.

## 2.5 Channel Classification

Normalized cues were organized into nine behavioral channels, summarized in Table 12. Facial expressions dominate the corpus, accounting for over a third of all mappings, followed by eye movements and body posture/movement. The relative scarcity of multimodal mappings reflects the historical tendency of studies to focus on a single channel rather than to integrate evidence across modalities.

**Table 12.** *Channel Distribution*

| Channel | Mappings | % | Detection Mode |
|---|---|---|---|
| Facial Expressions | 6,155 | 36.1% | Observable |
| Eye Movements | 2,439 | 14.3% | Mixed |
| Body Posture/Movement | 2,356 | 13.8% | Observable |
| Behavioral | 2,197 | 12.9% | Instrumental |
| Voice/Paralinguistic | 1,159 | 6.8% | Observable |
| Head Movements | 1,029 | 6.0% | Observable |
| Hand/Arm Gestures | 810 | 4.8% | Observable |
| Physiology | 801 | 4.7% | Instrumental |
| Multimodal | 93 | 0.5% | Instrumental |

## 2.6 Framework Construction

The processed data were organized into the four-level Nonverbal Syntax Framework presented in Section 3. At a high level, Level 1 (Cue Vocabulary) catalogs the 6,434 normalized cues organized by channel, with each cue annotated by its component evidence tier, paper count, associated states, and actionability level. Level 2 (State Clusters) inverts this view, providing for each of the 2,010 normalized states its associated cues, channel distribution, and top indicators. Level 3 (State Profiles) builds upon the clusters by constructing multimodal behavioral signatures for each state, with detailed profiles for the key learning states of engagement, confusion, frustration, boredom, and attention. Level 4 (Discriminative Analysis) identifies 1,215 state pairs that share at least three cues and computes Jaccard similarity to quantify confusion risk, then extracts the discriminative cues unique to each state in a pair, annotated with their relationship evidence tiers. The complete dataset, normalization dictionaries, and analysis scripts are available as Supplementary Materials.

# 3. THE NONVERBAL SYNTAX FRAMEWORK

The Nonverbal Syntax Framework organizes the synthesized, normalized, and tiered evidence into a four-level hierarchical structure designed to support both research and practical applications. This section presents each level in detail, illustrating how the framework enables systematic inference from observed cues to probable states.

## 3.1 Framework Overview

The framework adopts the metaphor of "nonverbal syntax" to capture how behavioral cues combine in patterned ways to signal internal states, analogous to how words combine according to grammatical rules to convey meaning. Just as understanding syntax enables language comprehension, understanding the nonverbal syntax of learning enables more accurate state inference.

The four levels build progressively, as summarized in Table 13.

**Table 13.** *Overview of the Four Levels of the Nonverbal Syntax Framework*

| Level | Name | Function | Primary Use Case |
|---|---|---|---|
| Level 1 | Cue Vocabulary | Catalog of all normalized cues | "What cues exist?" |
| Level 2 | State Clusters | Cues associated with each state | "What indicates state X?" |
| Level 3 | State Profiles | Multimodal signatures for states | "How does state X manifest?" |
| Level 4 | Discriminative Analysis | Distinguishing confusable states | "How to tell X from Y" |

Figure 2 illustrates the framework architecture and the relationships between levels.

## 3.2 Level 1: Cue Vocabulary

The Cue Vocabulary provides a comprehensive catalog of 6,434 normalized behavioral cues organized by channel. Each cue entry includes its evidence tier, the number of papers documenting it, and the states it has been associated with.

### 3.2.1 Channel Organization

Cues are organized into nine behavioral channels, reflecting distinct modalities of nonverbal expression, as shown in Table 14.

The "Detection" column distinguishes between directly observable cues (detectable by human observers without instrumentation), instrumental cues (requiring sensors or logging systems), and mixed channels (some variants observable, others requiring specialized equipment such as eye-tracking).

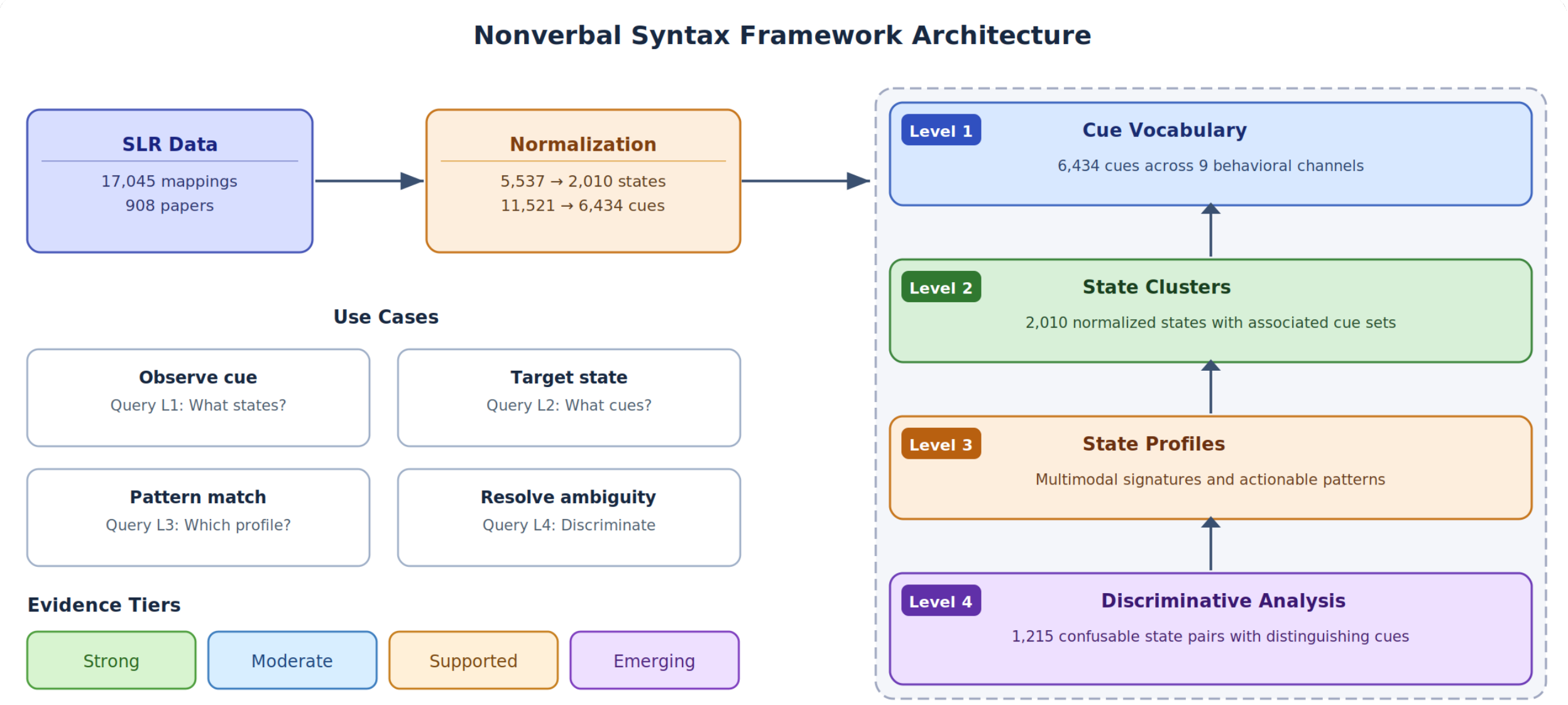

***Figure 2.*** *The Nonverbal Syntax Framework architecture. Data flows from the systematic review (17,043 mappings from 908 studies; see Turaev et al., 2026) through normalization, which reduces 5,537 raw state labels to 2,010 canonical states and 11,521 raw cue descriptions to 6,434 normalized cues. The normalized evidence is organized into a four-level hierarchical framework: Level 1 catalogs cues by behavioral channel; Level 2 clusters cues by state; Level 3 provides multimodal state profiles with detailed behavioral signatures; and Level 4 enables discrimination between confusable states through the identification of 1,215 state pairs and their distinguishing cues. Both component evidence tiers (how well-studied cues and states are individually) and relationship evidence tiers (how well-documented specific cue-state links are) are maintained throughout the framework. The "Use Cases" panel illustrates the four primary query patterns the framework supports: observing a cue and retrieving probable states (L1), targeting a state and retrieving its indicative cues (L2), pattern-matching observed behavior against state profiles (L3), and resolving ambiguity between confusable candidates (L4).*

**Table 14.** *Cue Counts and Detection Modes by Behavioral Channel*

| Channel | Cue Count | Example Cues | Detection |
|---|---|---|---|
| Behavioral | 1,668 | response time, mouse movements, interaction patterns | Instrumental |
| Facial Expressions | 1,581 | smile (125 papers), brow lowerer (69), frown (45) | Observable |
| Body Postures/Movements | 967 | body posture (251), slouching (10), leaning forward (23) | Observable |
| Eye Movements | 884 | eye gaze (189), fixation (83), blink (83) | Mixed |
| Voice/Paralinguistic | 408 | speech (148), voice (123), pitch (46), sighing (6) | Observable |
| Hand/Arm Gestures | 374 | gestures (156), hand raising (14), pointing (70) | Observable |
| Head Movements | 336 | head pose (171), head nod (55), head tilt (37) | Observable |
| Physiology | 177 | skin conductance (115), heart rate (98), EEG (67) | Instrumental |
| Multimodal | 39 | combined/fused features across channels | Instrumental |

### 3.2.2 Cue Evidence Tier Distribution

The component evidence tier distribution reveals that most cues are documented in only one study, as shown in Table 15.

**Table 15.** Cue Evidence Tier Distribution

| Tier | Criterion | Count | Percentage |
|---|---|---|---|
| Tier 1: Strong | ≥10 papers | 64 | 1.0% |
| Tier 2: Moderate | 5–9 papers | 45 | 0.7% |
| Tier 3: Supported | 3–4 papers | 92 | 1.4% |
| Tier 4: Emerging | 2 papers | 222 | 3.5% |
| Tier 5: Exploratory | 1 paper | 6,011 | 93.4% |

This distribution reflects the field's diversity; many specific cue variants have been studied only once, while highlighting a core set of 201 well-replicated cues (Tiers 1–3) that provide a reliable foundation for state inference.

Figure 3 illustrates the relationships among cues, states, and channels.

### 3.2.3 Usage Example

The Cue Vocabulary supports two complementary query patterns. A researcher or practitioner who has observed a particular behavior can look it up to discover which states it has been associated with in the literature, while one who is interested in a particular state can query the vocabulary to retrieve the full set of cues documented for that state, ranked by relationship evidence. The latter pattern is illustrated below using *confusion* as the target state. The query returns 542 cue-state relationships spanning eight of the nine behavioral channels, of which the top entries by relationship evidence are listed. Each entry includes the number of independent papers documenting the link and the corresponding relationship-evidence tier, enabling the user to immediately gauge how confidently each cue can be applied as a confusion indicator.

```
Query:  cues associated with "confusion"
Result: 542 cue-state relationships across 8 channels

Top specific cues by relationship evidence:
  AU4 (brow lowerer)               35 papers   R1: Strong
  AU7 (lid tightener)              14 papers   R2: Substantial
  AU12 (lip corner puller)         11 papers   R2: Substantial
  frown                             8 papers   R3: Moderate
  AU1 (inner brow raiser)           7 papers   R3: Moderate
  repeated fixation on elements     6 papers   R3: Moderate
  scratching head                   5 papers   R3: Moderate
```

The result illustrates the layered evidence structure that the framework enables: AU4 (brow lowerer) emerges as the single most replicated indicator of confusion, with thirty-five independent papers documenting the link (R1, Strong), while accompanying cues such as scratching head and repeated fixation provide moderate but converging support. A practitioner or system designer can use this output directly to select features for an automated detector, to construct an observation rubric, or to identify under-replicated cues that warrant further empirical investigation.

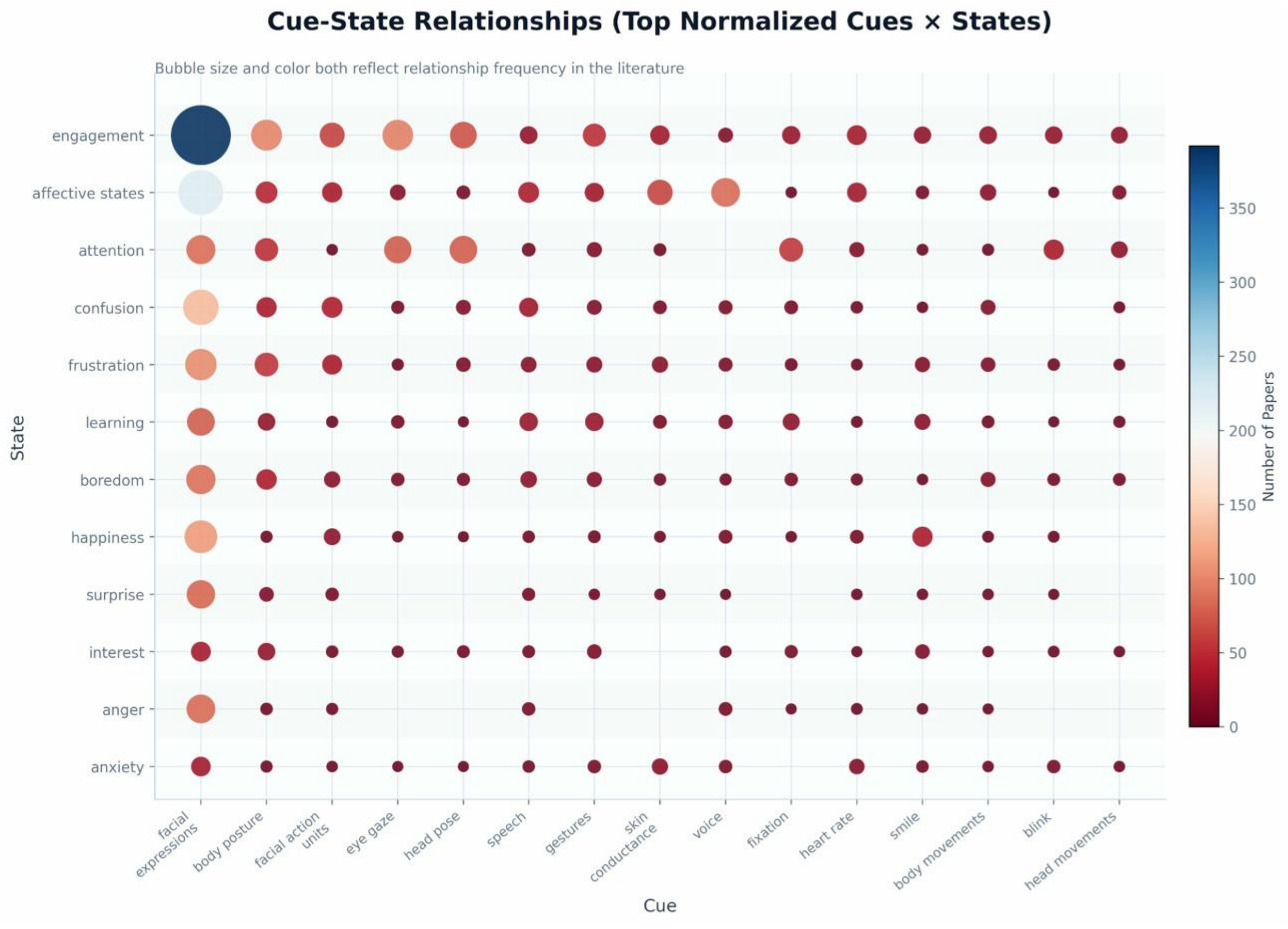

***Figure 3.*** *Cue-state relationship strength across the top normalized cues and states. The horizontal axis arranges the most extensively studied cues (facial expressions, body posture, facial action units, eye gaze, head pose, and so on) and the vertical axis arranges the most extensively studied states (engagement, affective states, attention, confusion, frustration, and others). Bubble size and color both encode the number of independent papers documenting each specific cue-state relationship, with larger darker bubbles indicating extensively replicated links. The visualization highlights the concentration of evidence for a small number of core cue-state combinations, most prominently facial expressions as indicators of engagement and affective states, while also illustrating the broad but sparse coverage of the remaining combinations.*

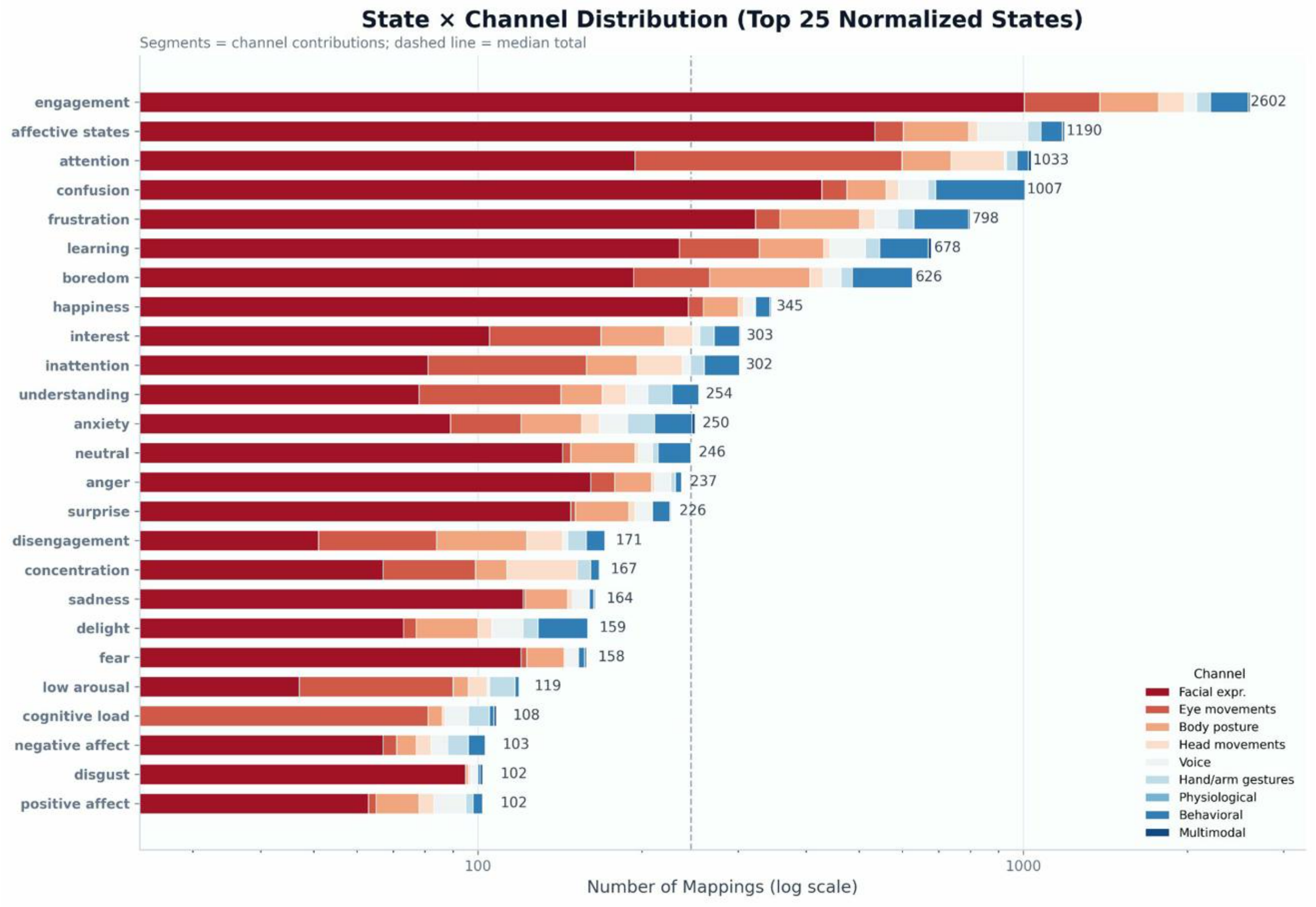

***Figure 4.*** *Channel distribution of the top 25 normalized states, displayed as a horizontal stacked bar chart on a log scale. Each bar represents one state, and the segments within each bar show the contribution of each of the nine behavioral channels to that state's total mapping count. Channels are color-coded from facial expressions (dark red) through multimodal features (dark blue), and the total number of mappings for each state is printed at the right end of its bar. The dashed vertical line indicates the median total across the displayed states. The visualization reveals that facial expressions dominate the evidence base for nearly all of the canonical affective states, that engagement is the only state with substantial evidence across all nine channels, and that eye movement and body posture cues contribute disproportionately to attention, confusion, and boredom respectively.*

### 3.3.2 Top States by Evidence

The most extensively studied states are summarized in Table 17.

**Table 17.** *Top 10 States by Number of Papers*

| Rank | State | Papers | Total Cues | Channels | R1–R4 Relationships |
|---|---|---|---|---|---|
| 1 | engagement | 471 | 1,307 | 9 | 81 |
| 2 | affective states (general) | 360 | 606 | 9 | 45 |
| 3 | confusion | 292 | 522 | 8 | 34 |
| 4 | attention | 281 | 500 | 9 | 33 |
| 5 | frustration | 261 | 401 | 9 | 33 |
| 6 | boredom | 242 | 335 | 8 | 30 |
| 7 | happiness | 202 | 134 | 9 | 15 |
| 8 | learning | 199 | 406 | 9 | 28 |
| 9 | surprise | 150 | 100 | 8 | 11 |
| 10 | anger | 148 | 116 | 9 | 12 |

The "R1–R4 Relationships" column indicates cue-state links documented by 3 or more independent papers, which is the actionable subset suitable for confident practical application.

Figure 4 presents the state-by-channel distribution for the top 25 states.

### 3.3.3 Cluster Structure

Each state cluster contains a cue count (the total cues associated with the state), a channel breakdown (cues organized by behavioral channel), a list of top cues with relationship evidence tiers, and the state's component evidence level. The cluster for *confusion* illustrates this structure:

```
State: confusion
Component Evidence: Tier 1 (Strong) — 292 papers
Total Cue Relationships: 542
Channels: 8 of 9

Channel Breakdown:
  Facial Expressions:       168 cues
  Eye Movements:             88 cues
  Body Posture/Movement:     72 cues
  Voice/Paralinguistic:      45 cues
  Behavioral:                74 cues
  Head Movements:            42 cues
```

```
  Hand/Arm Gestures:         35 cues
  Physiology:                18 cues

Top R1–R4 Cues (≥3 papers):
  AU4 (brow lowerer)             35 papers   R1
  AU7 (lid tightener)            14 papers   R2
  AU12 (lip corner puller)       11 papers   R2
  frown                           8 papers   R3
  AU1 (inner brow raiser)         7 papers   R3
  repeated fixation on elements   6 papers   R3
  scratching head                 5 papers   R3
  verbal: "I'm confused"          4 papers   R4
  head tilt (questioning)         4 papers   R4
```

## 3.4 Level 3: State Profiles

State Profiles provide multimodal behavioral signatures, that is, detailed specifications of how each state manifests across multiple channels, with emphasis on actionable, observable cues.

### 3.4.1 Profile Structure

Each profile includes a state summary (definition and educational relevance), a channel-by-channel breakdown of top cues with relationship evidence, a list of actionable indicators for direct detection, verbal indicators when documented, and discriminative cues that distinguish the state from confusable alternatives. Figure 5 visualizes the resulting channel-distribution profiles for the six most extensively studied learning states, showing how each state allocates its observable footprint across the seven major behavioral channels.

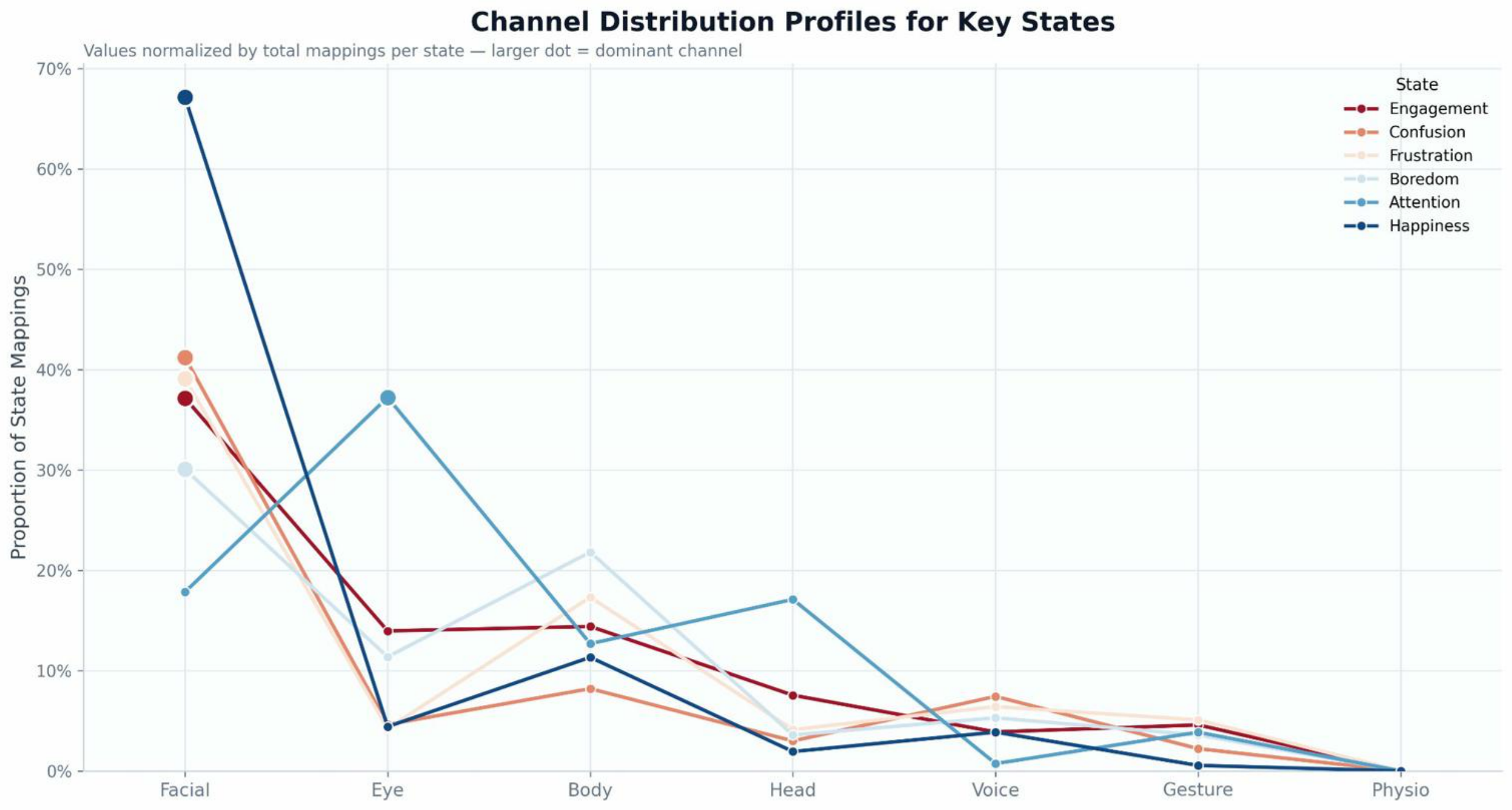

***Figure 5.*** *Channel distribution profiles for key learning states. Each line represents one state (engagement, confusion, frustration, boredom, attention, happiness) and traces the proportion of that state's total mappings across the seven major behavioral channels (facial expressions, eye movements, body posture, head movements, voice/paralinguistic, hand/arm gestures, and*

*physiology). Dot size encodes the dominant channel for each state. The visualization reveals clear state-specific signatures: happiness is overwhelmingly facial (67%), attention peaks on eye movements (37%), boredom shifts relatively toward body posture, and engagement, confusion, and frustration share a facial-dominated profile with secondary contributions across all remaining channels. These distinct channel profiles provide the basis for the multimodal state signatures presented in the detailed profiles of Sections 3.4.2–3.4.5.*

### 3.4.2 Example Profile: Confusion

**Definition.** A state of cognitive disequilibrium triggered by impasses, contradictions, or anomalies that conflict with expectations (D'Mello & Graesser, 2014).

**Educational Relevance.** Confusion can be productive when it triggers effortful processing that resolves the impasse, leading to deeper learning. However, prolonged or unresolved confusion is associated with frustration and disengagement.

**Multimodal Signature** (Table 18):

**Table 18.** Confusion Multimodal Signature

| Channel | Top Cues | Evidence |
|---|---|---|
| Facial | AU4 brow lowerer (35), AU7 lid tightener (14), AU12 lip corner puller (11), frown (8) | R1–R3 |
| Eye | Repeated fixation on same elements (6), gaze toward material (5), increased blink rate (4) | R3–R4 |
| Head | Head tilt (questioning) (4), head shake (3) | R4 |
| Body | Leaning toward screen (3), stillness/pause (3) | R4 |
| Gesture | Scratching head (5), self-touch (3), hand to chin (3) | R3–R4 |
| Voice | Verbal: "I don't understand" (4), questioning intonation (3), "Why?" (3) | R4 |

**Actionable Observable Indicators** (for human observers): furrowed brow (inner brows drawn together); squinting or narrowed eyes; scratching or touching the head; tilted head with questioning expression; gaze fixed on the problem area; verbal expressions of not understanding.

**Verbal Self-Reports** (high confidence when present): "I'm confused.", "I don't get it.", "Why didn't it work?", "This doesn't make sense."

### 3.4.3 Example Profile: Frustration

**Definition.** A negative affective state arising from blocked goals, repeated failures, or perceived lack of progress (Kapoor et al., 2007).

**Educational Relevance.** Unlike productive confusion, frustration is generally associated with negative outcomes, such as reduced persistence, surface-level strategies, and eventual disengagement.

**Multimodal Signature** (Table 19):

**Table 19.** Frustration Multimodal Signature

| Channel | Top Cues | Evidence |
|---|---|---|
| Facial | AU4 brow lowerer (15), frown (12), tightened jaw (8), AU23 lip tightener (6) | R2–R3 |
| Eye | Gaze away from task (7), eye rolling (3) | R3–R4 |
| Head | Head shake (negative) (4), head drop (3) | R4 |
| Body | Tense posture (5), restlessness (4), leaning back (3) | R3–R4 |
| Gesture | Banging on keyboard (5), pulling hair (4), banging on mouse (4), clenched fists (3) | R3–R4 |
| Voice | Sighing/deep sighing (6), raised voice (4), groaning (3), verbal: "This is stupid" (3) | R3–R4 |

**Actionable Observable Indicators:** deep sighing or groaning; aggressive actions (banging keyboard/mouse); hair pulling or self-directed aggression; tense, rigid posture; raised voice or exclamations; head shaking or head drop.

**Verbal Self-Reports:** "This is frustrating.", "I give up.", "This is stupid.", expletives or exclamations of annoyance.

### 3.4.4 Example Profile: Boredom

**Definition.** A state of low arousal and disengagement characterized by lack of interest and perceived meaninglessness of the current activity (Pekrun et al., 2010).

**Educational Relevance.** Boredom predicts off-task behavior, superficial processing, and poor learning outcomes. Early detection enables intervention through task variation or difficulty adjustment.

**Multimodal Signature** (Table 20):

**Table 20.** Boredom Multimodal Signature

| Channel | Top Cues | Evidence |
|---|---|---|
| Facial | Neutral/flat expression (12), yawning (8), drooping eyelids (5) | R2–R3 |
| Eye | Gaze wandering away (9), looking at clock/door (4), reduced fixation (4) | R3–R4 |
| Head | Head resting on hand/palm (4), head propping (3) | R4 |
| Body | Slouching (10), slumped posture (6), resting chin on palm (4) | R2–R4 |
| Gesture | Fidgeting (7), doodling (3), playing with objects (3) | R3–R4 |
| Voice | Monotone voice (3), verbal: "This is boring" (4) | R4 |

**Actionable Observable Indicators:** slouching or slumped posture; yawning; resting chin or head on palm; gaze wandering away from the task; fidgeting or doodling; flat, expressionless face.

**Verbal Self-Reports:** "This is boring.", "When will this be over?", heavy sighing (different from frustration sighing, which is less forceful).

### 3.4.5 Example Profile: Engagement

**Definition.** A state of focused involvement, interest, and active participation in the learning activity (Fredricks et al., 2004).

**Educational Relevance.** Engagement is the most extensively studied state (471 papers) and is consistently associated with positive learning outcomes, persistence, and deeper processing.

**Multimodal Signature** (Table 21):

**Table 21.** Engagement Multimodal Signature

| Channel | Top Cues | Evidence |
|---|---|---|
| Facial | Smile (18), raised eyebrows (interest) (9), attentive expression (7) | R2–R3 |
| Eye | Eye contact with material (12), focused gaze (10), reduced blinking (6) | R2–R3 |
| Head | Head nodding (16), upright head position (5), head orientation toward task (4) | R2–R4 |
| Body | Forward lean (9), upright posture (7), oriented toward task (6) | R3 |
| Gesture | Taking notes (8), hand raising (6), gesturing while explaining (4), asking questions (3) | R3–R4 |
| Voice | Active verbal participation (8), questions (5), discussion (4) | R3–R4 |

**Actionable Observable Indicators:** forward lean toward the material; head nodding during instruction; taking notes; hand raising to ask or answer questions; focused gaze on the speaker or the material; smiling or animated expression.

**Verbal Indicators:** asking questions; contributing to the discussion; verbal expressions of interest.

## 3.5 Level 4: Discriminative Analysis

The Discriminative Analysis level addresses the *state-ambiguity* problem, the challenge that many cues are associated with multiple states, leading to inference uncertainty.

### 3.5.1 Identifying Confusable State Pairs

We identified 1,215 state pairs that share at least three cues, indicating potential confusion risk. For each pair, we computed the shared cues (cues documented for both states), the state-specific cues (cues documented for only one state in the pair), and the Jaccard similarity (an overlap coefficient quantifying confusion risk). The most confusable pairs are presented in Table 22. Figure 6 visualizes the top pairs in two complementary views: the overall highest-Jaccard pairs in the full 1,215-pair set, and the subset of top pairs that involve at least one of the canonical learning states. These two views together make the structure of confusion risk in the framework immediately apparent: the highest absolute Jaccard values occur among affective-dimensional pairs (e.g., arousal and valence, empathy and rapport), while the most practically consequential confusions for educational applications occur among the learning-related pairs such as confusion-frustration and boredom-frustration.

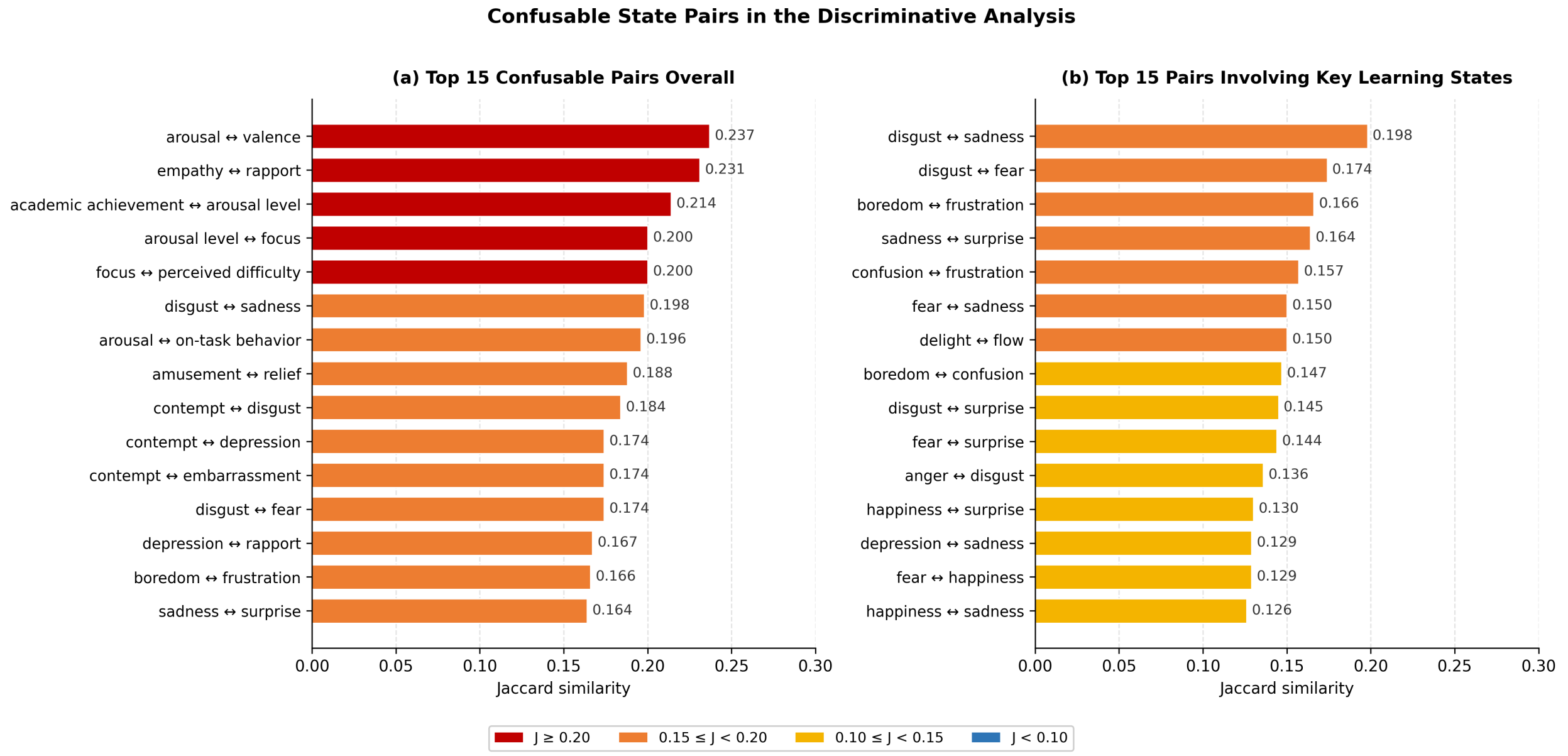

***Figure 6.*** *Top confusable state pairs ranked by Jaccard similarity. Panel (a) shows the fifteen highest-similarity pairs in the full set of 1,215 confusable state pairs. Panel (b) restricts the same ranking to pairs involving at least one canonical learning state (engagement, confusion, frustration, boredom, attention, learning, happiness, anxiety, interest, cognitive load, delight, concentration, disengagement, inattention, surprise, anger, sadness, fear). Bar color encodes the Jaccard similarity level, with red bars indicating high confusion risk ($J \geq 0.20$), orange moderate risk ($0.15 \leq J < 0.20$), yellow lower risk ($0.10 \leq J < 0.15$), and blue minimal risk ($J < 0.10$). The highest Jaccard value observed in the entire framework is 0.237 (arousal versus valence), and no pair exceeds 0.25, indicating that even the most confusable states in the framework retain substantial discriminative evidence that can be recovered through the cue-level analysis of Section 3.5.2.*

**Table 22.** Selected Confusable State Pairs Involving Key Learning States

| State 1 | State 2 | Shared Cues | Jaccard |
|---|---|---|---|
| confusion | frustration | 125 | 0.157 |
| boredom | frustration | — | 0.166 |
| boredom | confusion | — | 0.147 |
| engagement | attention | 76 | 0.044 |
| boredom | disengagement | 20 | 0.048 |
| frustration | anger | 21 | 0.042 |
| confusion | cognitive load | 18 | 0.032 |

The Jaccard similarity values in this table are computed from the full normalized cue sets for each state, without filtering by channel or paper count. Higher Jaccard similarity indicates greater confusion risk; the maximum value observed in the full 1,215-pair set is 0.237 (arousal versus valence), and no pair exceeds 0.25, indicating that while many pairs share a substantial number of cues, each state also retains a set of unique indicators sufficient for discrimination.

### 3.5.2 Discriminative Cue Analysis

For each confusable pair, we identify cues unique to each state with their relationship evidence. Tables 23, 24, and 25 present the discriminative cues for three illustrative pairs that span the most extensively studied learning states.

**Table 23.** Confusion vs. Frustration: Discriminative Cues

| **Confusion-Specific Cues** | **Evidence** | **Frustration-Specific Cues** | **Evidence** |
|---|---|---|---|
| scratching head | 5 papers (R3) | sighing / deep sighing | 6 papers (R3) |
| head tilt (questioning) | 4 papers (R4) | banging on keyboard | 5 papers (R3) |
| “Why didn’t it work?” | 3 papers (R4) | pulling hair | 4 papers (R4) |
| gaze toward material | 3 papers (R4) | banging on mouse | 4 papers (R4) |
| looking at classmate’s work | 3 papers (R4) | clenched jaw / raised voice | 4 papers (R4) |

**Table 24.** Boredom vs. Confusion: Discriminative Cues

| **Boredom-Specific Cues** | **Evidence** | **Confusion-Specific Cues** | **Evidence** |
|---|---|---|---|
| slouching | 10 papers (R2) | furrowed brow / AU4 | 35 papers (R1) |
| yawning | 8 papers (R3) | scratching head | 5 papers (R3) |
| resting chin on palm | 4 papers (R4) | gaze toward material | 3 papers (R4) |
| gaze wandering away | 9 papers (R3) | repeated fixation | 6 papers (R3) |
| decreased activity | 4 papers (R4) | increased self-touch | 3 papers (R4) |

**Table 25.** Engagement vs. Attention: Discriminative Cues

| Engagement-Specific Cues | Evidence | Attention-Specific Cues | Evidence |
|---|---|---|---|
| head nodding | 16 papers (R2) | leaning backward | 6 papers (R3) |
| taking notes | 8 papers (R3) | supporting head | 3 papers (R4) |
| hand raising | 6 papers (R3) | passive gaze | 3 papers (R4) |
| asking questions | 3 papers (R4) | — | — |
| smile | 18 papers (R2) | — | — |

### 3.5.3 Diagnostic Application

When a practitioner observes behaviors consistent with multiple states, discriminative cues resolve ambiguity. Two illustrative scenarios show how this works in practice.

*Scenario 1.* A student displays a furrowed brow and reduced eye contact. Both confusion and frustration include these cues. Discriminative analysis suggests: if scratching head is also present, the most likely state is confusion; if sighing is present, the most likely state is frustration; if banging on the keyboard is present, the most likely state is also frustration; and if "Why didn't it work?" is verbalized, the most likely state is confusion.

*Scenario 2.* A student shows reduced activity and occasional yawning. Both boredom and fatigue include these cues. Discriminative analysis suggests: if slouching and resting chin on palm are also present, the most likely state is boredom; if eye rubbing and stretching are present, the most likely state is fatigue.

## 3.6 Framework Integration

The four levels work together to support inference. The user observes a cue and queries Level 1 for associated states; identifies candidate states and queries Level 2 for additional expected cues; compares the observed pattern against Level 3 profiles; and resolves any remaining ambiguity by querying Level 4 for discriminative cues between candidates.

### 3.6.1 Worked Example

*Observation.* A student shows a furrowed brow, repeated looks at the same interface element, and scratches their head.

*Step 1 (Level 1).* Each cue maps to multiple states: furrowed brow → confusion (35), frustration (15), engagement (13); repeated looks at the same element → confusion (6); scratching head → confusion (5).

*Step 2 (Level 2).* Confusion appears in all three mappings; thus, check its cluster. The confusion cluster includes all three observed cues, and the pattern is highly consistent with confusion.

*Step 3 (Level 3).* Compare with confusion profile: facial (furrowed brow) at R1 Strong evidence; eye (repeated looks) at R3 Moderate evidence; gesture (scratching head) at R3 Moderate evidence — three high-evidence cues converging on confusion.

*Step 4 (Level 4).* Verify against confusable states: confusion versus frustration shows no frustration discriminators present (no sighing, banging, hair pulling), while a confusion discriminator (scratching head) is present.

*Inference.* High confidence for confusion. The multi-cue pattern (furrowed brow + repeated fixation + scratching head) provides stronger evidence than any single cue alone. This is the *nonverbal syntax* principle: the combination forms a reliable "sentence" indicating confusion.

### 3.6.2 Worked Example: Mixed State

*Observation.* A student shows a furrowed brow, a forward lean, and head nodding.

*Step 1 (Level 1).* Each cue maps to states: furrowed brow → confusion (35), frustration (15), engagement (13); forward lean → engagement (9), attention (3); head nodding → engagement (16), attention (4).

*Step 2 (Level 2).* No single state includes all three as top cues: confusion includes furrowed brow but not forward lean or nodding; engagement includes forward lean and nodding, but furrowed brow is lower-ranked.

*Step 3 (Level 3).* Compare profiles: the confusion profile features furrowed brow but typically shows leaning back, not forward; the engagement profile features forward lean and head nodding, but a furrowed brow within an engagement context indicates concentration.

*Step 4 (Level 4).* Consider mixed state: forward lean and head nodding are engagement discriminators, while a furrowed brow in an engagement context suggests concentration or effort.

*Inference.* The combination suggests *engaged confusion* (or "productive confusion"), a state in which the student is actively working through a difficulty. D'Mello and Graesser (2012) document this as a common transitional state during learning. The presence of engagement indicators (forward lean, nodding) combined with cognitive effort indicators (furrowed brow) suggests the confusion is productive rather than overwhelming.

The Nonverbal Syntax Framework provides a comprehensive, evidence-graded system for inferring learner states from observable behavior. Level 1 contributes a cue catalog of 6,434 cues organized by nine channels with evidence tiers. Level 2 contributes a state-centric view of 2,010 states with cue clusters and channel distributions. Level 3 contributes multimodal signatures with detailed actionable cue specifications. Level 4 contributes disambiguation tools through discriminative cues for 1,215 confusable pairs. Together, these four levels identify 480 actionable relationships (R1–R4 evidence, ≥3 papers) covering 35.5% of all documented mappings across 47 states, 111 cues, and 8 of 9 channels. These represent the consolidated, replicated core of six decades of research that practitioners can apply with confidence, while the full corpus of 10,553 documented cue-state links remains available for comprehensive research applications.

The core insight, the *nonverbal syntax principle*, is that multi-cue patterns provide more reliable state indicators than individual cues alone. A furrowed brow is ambiguous; a furrowed brow combined with scratching head and repeated fixation reliably indicates confusion. By documenting these patterns with explicit evidence tiers and actionability classifications, the framework transforms fragmented research findings into an integrated, usable resource for understanding and inferring learner states from observable behavior.

# 4. DISCUSSION

The Nonverbal Syntax Framework represents a systematic effort to transform fragmented research findings into an integrated, evidence-graded resource for understanding and inferring learner states from observable behavior. This section discusses the framework's contributions, implications, limitations, and future directions.

## 4.1 Summary of Contributions

### 4.1.1 Terminological Consolidation

A fundamental contribution is the development of comprehensive normalization procedures that address the field's terminological fragmentation. The reduction from 5,537 raw state labels to 2,010 normalized states (63.7%) and from 11,521 cue descriptions to 6,434 normalized cues (44.2%) enables meaningful synthesis across studies that would otherwise be isolated by inconsistent terminology.

This consolidation reveals that apparent diversity in findings often reflects labeling differences rather than substantive disagreement. For example, the 666 variants of "engagement" can now be recognized as referring to a common construct, strengthening evidence for engagement-related cue associations from scattered single-study observations to a robust multi-study foundation.

### 4.1.2 Dual-Evidence Assessment

The framework introduces a novel evidence system operating at two complementary levels. *Component Evidence Tiers* assess how well-studied individual cues and states are across all their associations, while *Relationship Evidence Tiers* (R1–R6) assess how well-documented each specific cue-state pair is. Tables 26 and 27 summarize the two distributions side by side.

**Table 26.** Component Evidence Tier Distribution (States and Cues)

| Category | States | Cues | Interpretation |
|---|---|---|---|
| Tier 1: Strong | 34 | 64 | Core constructs with extensive replication |
| Tier 2: Moderate | 8 | 45 | Reliable with solid support |
| Tier 3: Supported | 21 | 92 | Reasonable empirical backing |
| Tier 4: Emerging | 118 | 222 | Preliminary findings requiring validation |
| Tier 5: Exploratory | 1,829 | 6,011 | Single-study observations |

**Table 27.** Relationship Evidence Tier Distribution

| Tier | Criterion | Count | Interpretation |
|---|---|---|---|
| R1: Strong | ≥20 papers | 47 | Extensively replicated link |
| R2: Substantial | 10–19 papers | 63 | Well-documented relationship |
| R3: Moderate | 5–9 papers | 137 | Solid replication |
| R4: Supported | 3–4 papers | 233 | Reasonable evidence |
| R5: Emerging | 2 papers | 420 | Preliminary, needs validation |
| R6: Exploratory | 1 paper | 9,653 | Single-study hypothesis |

This dual assessment reveals a critical finding: 52% of "Very High" combined confidence relationships are documented by only a single paper for that specific link. A cue and state may both be extensively researched individually, yet their specific combination may have minimal direct evidence. The framework makes this distinction explicit, enabling practitioners to make appropriately calibrated inferences.

#### 4.1.3 Contextualizing Actionable Relationships

The framework identifies 480 actionable relationships with R1–R4 evidence (≥3 papers), representing only 4.5% of unique cue-state pairs but covering 35.5% of all documented mappings (6,049 of 17,043). These relationships span 47 states (including all key learning states), 111 distinct cues (observable behavioral indicators), and 8 of 9 channels.

This distribution exhibits the long-tailed structure expected of cumulative scientific evidence. To characterize it formally, we fit a discrete power-law to the relationship paper-count distribution using the maximum-likelihood method of Clauset, Shalizi, and Newman (2009), as implemented in the *powerlaw* Python package (Alstott, Bullmore, & Plenz, 2014). The fit yields a scaling exponent of $\alpha = 2.13$ (SE = 0.05) above a data-driven lower cutoff of $x_{\min} = 3$ papers, which is well within the 2–3 range typical of empirical power-law distributions in scientific and citation contexts. A bootstrap goodness-of-fit test based on 1,000 synthetic datasets returned $p \approx 1.0$ (Kolmogorov–Smirnov distance $D = 0.022$), indicating that the power-law hypothesis cannot be rejected. Likelihood-ratio comparisons further showed that the power-law was strongly preferred over an exponential alternative ($R = +5.66$, $p < 0.001$), while comparisons against lognormal and stretched-exponential alternatives were statistically inconclusive ($p = 0.29$ and $p = 0.53$, respectively), as is common for heavy-tailed empirical distributions of this size. Figure 7 displays the empirical complementary CDF and probability density on log-log axes alongside the fitted power-law.

A striking feature of the maximum-likelihood fit is that the data-driven lower cutoff $x_{\min} = 3$ coincides exactly with the framework's a priori threshold for actionability. The R1–R4 subset, defined methodologically as relationships documented by three or more independent papers, is also empirically the regime in which the relationship paper-count distribution becomes well described by a power-law. The 480 actionable relationships therefore correspond to the head of a long-tailed distribution that has been formally characterized rather than assumed, and the consolidated, replicated core of the framework sits alongside an exploratory periphery of 9,653 single-study findings whose evidential status is qualitatively distinct.

#### 4.1.4 Actionability Classification

The framework distinguishes between cues based on their practical observability, with 12,882 mappings (52%) involving highly actionable cues, 3,908 (16%) moderately actionable, 253 (1%) weakly actionable, and 7,666 (31%) non-actionable. This classification bridges the gap between research findings and practical application. While "facial expressions" may be documented in hundreds of papers, a practitioner cannot observe “facial expressions” directly; they must observe specific behaviors like “furrowed brow” or “smile.” Over two-thirds of mappings (67.9%) involve cues that are highly or moderately actionable, enabling their application in observation-based contexts.

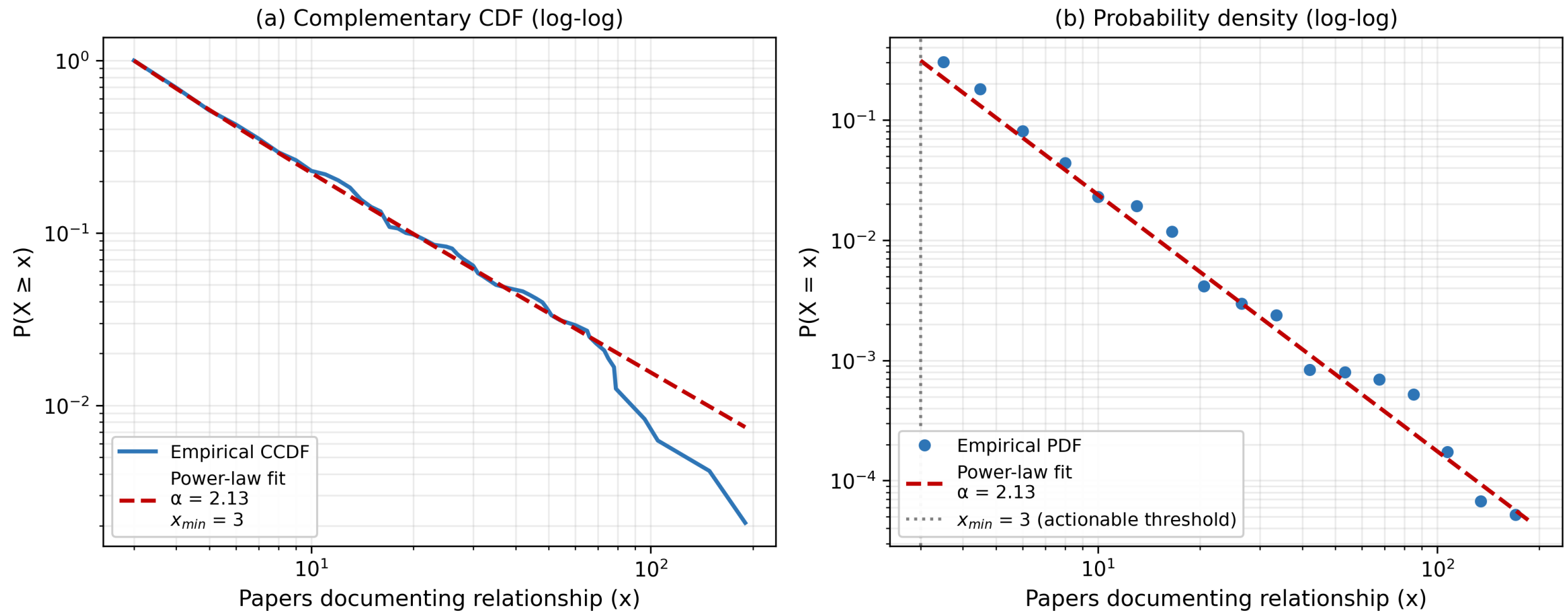

***Figure 7.*** *Power-law fit to the distribution of papers per cue-state relationship. (a) Complementary cumulative distribution function on log-log axes, with the empirical CCDF (solid blue) and the maximum-likelihood power-law fit (dashed red). (b) Probability density function on log-log axes, with empirical points (blue) and fitted power-law (dashed red). The data-driven lower cutoff $x_{min}$ = 3 (vertical dotted line) coincides with the framework's a priori threshold for actionable relationships (R1–R4, ≥3 papers). The fit yields α = 2.13 with goodness-of-fit p ≈ 1.0 (N = 10,557 relationships).*

### 4.1.5 Observable versus Instrumental Distinction

The framework explicitly distinguishes between directly observable channels (Facial, Head, Body, Gesture, Voice, and Eye in part), which are accessible to educators conducting real-time observation, and instrumentally measured channels (Physiology, Behavioral, Multimodal, and the detailed components of Eye), which require sensors or logging infrastructure typically available only in automated systems and research laboratories. This distinction enables practitioners to filter the framework to cues appropriate for their detection context, addressing the common concern that research findings often require instrumentation unavailable in typical educational settings.

### 4.1.6 Structured Organization

The four-level hierarchical structure serves distinct use cases. Level 1 (Cue Vocabulary) supports cue-first exploration, answering “What might this behavior indicate?” Level 2 (State Clusters) supports state-first detection, answering “What should I look for to detect X?” Level 3 (State Profiles) supports pattern matching, answering “Does this pattern match state X?” Level 4 (Discriminative Analysis) supports disambiguation, answering “How to tell X from Y?” This multi-entry organization supports both research (systematic evidence lookup) and practice (real-time inference support).

### 4.1.7 Discriminative Analysis

Perhaps the most practically valuable contribution is the discriminative analysis of confusable state pairs. The identification of 1,215 state pairs sharing at least three cues and the specification of discriminative cues for each pair address a critical gap in existing resources. While prior work has documented what cues indicate a given state, the framework uniquely addresses what cues distinguish between states that share common indicators. For example, both confusion and frustration involve furrowed brow (AU4), but the framework identifies that *sighing* (6 papers), *banging on keyboard* (5 papers), and *pulling hair* (4 papers) are discriminative for frustration, while *scratching head* (5 papers) and verbal statements like "Why didn't it work?" (3 papers) are discriminative for confusion.

## 4.2 Implications for Research

### 4.2.1 Identifying Knowledge Gaps

The tier distributions reveal striking imbalances in the evidence base. While a small set of core states (engagement, confusion, frustration, boredom) and cues (facial expressions, eye gaze, body posture) have been extensively studied, the vast majority of states and cues have been examined in only one or two studies. Priority areas for future research therefore include under-studied states such as curiosity, delight, and flow, which have theoretical importance but limited empirical documentation (Tier 3–4); under-studied channels such as gesture (374 cues) and voice (408 cues), which are underrepresented relative to facial (1,581 cues) and body (967 cues) channels; relationship replication, since of 10,553 documented cue-state relationships only 480 (4.5%) have R1–R4 evidence and the field would benefit from replication studies targeting theoretically important relationships currently at R5–R6; discriminative validation, since the discriminative cues identified here are derived analytically and direct comparison studies examining multiple states simultaneously are needed; and reporting of specific actionable cues, since many findings report generic cues (e.g., "facial expressions") without specifying observable behaviors that future studies should report.

### 4.2.2 Methodological Standardization

The normalization dictionaries provide a foundation for methodological standardization. Future studies can use canonical state and cue labels to ensure comparability, report results at multiple specificity levels (e.g., both "AU4" and "furrowed brow"), explicitly address discriminability from related states, classify cues by actionability level, and report the relationship evidence tier for claimed associations.

### 4.2.3 Theory Development

The framework supports theory development by revealing the empirical structure of cue-state relationships. Some states have concentrated channel profiles (boredom, dominated by body posture indicators), while others are truly multimodal (engagement, with documented cues across all nine channels). The skewed distribution of evidence suggests core relationships that should anchor theory, and the pattern of which cues discriminate which states informs theoretical models of state differentiation.

### 4.2.4 Complementarity with Deep Learning Approaches

The Nonverbal Syntax Framework is positioned as complementary to, rather than competitive with, contemporary deep-learning approaches to learner state detection. Deep neural networks have achieved impressive accuracy on benchmark facial expression and engagement datasets (Li & Deng, 2022; Whitehill et al., 2014), yet they typically operate as opaque mappings from raw pixels or sensor signals to state labels, are brittle when transferred across populations, recording conditions, or learning contexts (Kołakowska et al., 2017), and require large quantities of annotated training data that are expensive to collect in educational settings. The framework addresses these limitations along three complementary axes. First, the dual-evidence tiers and discriminative analyses provide empirically grounded *priors* for hybrid systems in which deep models are constrained by literature-derived knowledge of which cues should be informative for which states. Second, the actionable, observable cues catalogued in the framework constitute a validated *feature set* for explainable models, supporting interpretable architectures (e.g., concept-bottleneck models or rule-based components attached to learned representations) in which predictions can be traced back to specific behavioral indicators with explicit evidence tiers. Third, the framework provides a *ground-truth scaffold* for training-data annotation: when human coders or weakly supervised pipelines label educational video for state recognition, they can use the multimodal signatures and discriminative cues as a calibrated coding rubric, reducing annotation noise and ensuring that labels rest on the same evidence base across studies. In all three roles, the framework serves not as a substitute for high-capacity learned models but as the interpretable, evidence-graded foundation on which such models can be built and audited.

## 4.3 Implications for Practice

### 4.3.1 Evidence-Calibrated Observation

The framework enables practitioners to calibrate their confidence based on explicit evidence rather than intuition. Recommended thresholds depend on context: for high-stakes uses such as automated assessment or intervention triggers, only R1–R2 relationships (≥10 papers) should be relied upon, in order to minimize false positives. For general practice such as classroom observation or system design, the R1–R4 actionable subset (≥3 papers) provides a reasonable balance between coverage and confidence. For research exploration and hypothesis generation, the full corpus including R5–R6 relationships maximizes discovery potential at the cost of replication confidence.

### 4.3.2 Practical Behavioral Signatures

The detailed state profiles translate research into observable indicators. *Confusion* presents a distinctive facial pattern (furrowed brow, 35 papers) combined with self-directed gestures (scratching head, 5 papers) and fixation patterns (repeated looks at the same element, 6 papers). *Boredom* presents distinctive postural indicators (slouching, 10 papers; resting chin on palm, 4 papers) and facial indicators (yawning, 8 papers) that enable detection without sophisticated sensing. *Frustration versus confusion* can be distinguished via the voice channel (sighing) and gesture channel (banging on keyboard, pulling hair), enabling differentiated intervention between unproductive frustration and potentially productive confusion. *Engagement* requires a broad multimodal profile (9 channels) and integrated monitoring; the presence of participatory behaviors (hand raising, taking notes) distinguishes active engagement from passive attention.

## 4.4 Limitations

### 4.4.1 Data Foundation Limitations

The framework is built upon a systematic literature review with inherent limitations detailed in Turaev et al. (2026). The synthesized studies vary substantially in methodology, population, context, and measurement approach; while normalization addresses terminological differences, it cannot resolve deeper heterogeneity in operationalization. The evidence base reflects published research, which may over-represent positive findings and under-represent null results, so some cue-state relationships may appear more robust than warranted by the underlying phenomena. Most studies were conducted in educational or laboratory settings, and generalization to other contexts (such as workplace training or clinical assessment) requires further validation.

#### 4.4.2 Normalization Limitations

Normalization necessarily collapses distinctions that may be meaningful: "cognitive engagement" and "behavioral engagement" are mapped to a single "engagement" category, losing their theoretical distinction. Despite systematic procedures, some normalization decisions involve judgment, and the provided dictionaries enable researchers to examine and adjust mappings as needed. New terminology will continue to emerge that is not captured in current dictionaries, and the framework therefore requires ongoing maintenance.

#### 4.4.3 Evidence System Limitations

Users may confuse "Very High combined confidence" (well-studied components) with strong relationship evidence. The framework explicitly separates these, but careful interpretation is required.

The boundaries between evidence tiers (≥20, 10–19, 5–9, 3–4, 2, and 1 papers) are necessarily discrete and could in principle be drawn elsewhere. Rather than treating these as arbitrary, however, we adopted them on three principled grounds. First, the cut-offs correspond to commonly used thresholds in evidence-synthesis and meta-analytic practice, where a single observation, two observations, three to four observations, and five or more observations are typically taken to mark qualitatively distinct levels of replication strength. Second, the cut-offs separate the empirical regime in which a finding is meaningfully replicated (≥3 papers) from the regime in which it constitutes an isolated observation, and this is the practically relevant boundary for the framework's actionability claims. Third, because the underlying paper counts are openly available in the supplementary materials, users who prefer different thresholds can readily perform sensitivity analyses by reapplying alternative cut-offs to the raw data. We therefore present the tiers as principled guidelines rather than as hard, immutable cut-offs, and we encourage downstream users to treat them accordingly.

A second limitation of the present evidence system is that all papers are weighted equally. A study based on twelve students in a single laboratory contributes the same single increment to a relationship's paper count as a multi-site replication with five hundred participants, and a methodologically rigorous experimental design contributes no more than an exploratory observational report. This is a deliberate methodological choice rather than an oversight: the systematic literature review underlying the framework did not extract sample sizes, effect sizes, or methodological-quality scores uniformly across the 908 included papers, and any post-hoc weighting scheme imposed without that information would introduce more noise than it removed. The framework therefore reports replication *count* rather than aggregated *strength* of evidence, and we explicitly flag this as a limitation that future versions of the framework should address. A natural next step is to incorporate sample-size-weighted or quality-weighted evidence tiers, drawing on the established meta-analytic literature on study weighting, and to revisit the actionable subset under such a weighted scheme to determine how robust the current 480 actionable relationships are to changes in the weighting rule.

A third consideration concerns the *context dependence* of the synthesized relationships. The framework treats cue-state mappings as if they were context-invariant, yet some relationships are likely to be culturally or technologically robust (the link between smiling and positive affect, for example) while others are clearly tied to a specific learning context (hand-raising as an engagement indicator presupposes a classroom format that affords it; keystroke or mouse-movement cues presuppose a screen-based learning environment; head orientation toward the front of the room presupposes a teacher-fronted instructional setting). The framework currently does not annotate cues with context-sensitivity flags, and as a result a downstream user must apply judgment when transferring an actionable relationship from the context in which it was originally documented to a new setting. Adding explicit context-dependence annotations to the cue vocabulary is a priority for future versions, and would enable systematic filtering of the framework by deployment context (in-person vs. online, individual vs. classroom, teacher-led vs. self-paced).

Specific cues inferred from generic categories (e.g., "furrowed brow" from "facial expressions") may not perfectly reflect original study intent.

#### 4.4.4 Framework Limitations

The framework synthesizes correlational evidence; it does not establish causation. A cue associated with a state may not cause or definitively indicate that state. Furthermore, the framework describes population-level patterns: individual expressivity, cultural display rules, and contextual factors may produce substantial within-person and between-group variation. States are also treated as static categories, but real-world states fluctuate, blend, and transition, and the framework therefore provides snapshots rather than dynamic models.

#### 4.4.5 Temporal Validity and the Evolving Learning Landscape

A specific concern that warrants explicit treatment is the temporal validity of the synthesized relationships in light of the rapid evolution of educational technology and student behavior over the past five years. Two developments are particularly consequential. First, the COVID-19 pandemic accelerated a shift toward remote, hybrid, and screen-mediated instruction that reshaped both the contexts in which learner states must be observed and the behavioral repertoires through which those states are expressed. Second, the rapid integration of generative AI tools into learning environments has begun to alter how students engage with material, how they seek help, and how they manifest difficulty, with consequences for the observable footprint of states such as confusion, engagement, and frustration that are not yet fully understood. A reasonable concern is therefore whether cue-state relationships established in earlier classroom-based or pre-pandemic studies remain valid in contemporary digital settings.

Three considerations bear on this concern. First, the underlying behavioral channels of the framework, that is, facial action, gaze, posture, gesture, and prosody, are anchored in human physiology and emotional expression, and there is no reason to expect that the basic affective and cognitive expressions decoded by these channels have changed materially over the period covered by the present synthesis. A furrowed brow indicates cognitive disequilibrium in 2025 for the same reasons it did in 1985. Second, the empirical evidence base itself is heavily weighted toward recent research: 56.8% of the 908 studies underlying the framework were published between 2020 and 2025, ensuring that contemporary learning contexts, including remote and hybrid settings, are well represented in the actionable relationships. Third, however, certain context-bound cues are clearly products of, or transformed by, the digital era. Screen-oriented postures, fixation patterns specific to interface elements, and interaction-log behaviors such as keystroke dynamics or mouse-movement patterns have emerged or gained prominence only in recent years and are not represented in the older portions of the corpus. Conversely, traditional classroom indicators such as hand-raising or front-facing head orientation may be diminishing in relevance as instruction shifts away from teacher-fronted, in-person formats.

The net effect is that the framework's *core* relationships, those that depend on physiologically anchored channels and that are well represented in the recent literature, are likely to remain valid across the contemporary learning landscape, while a smaller set of context-bound relationships will require explicit re-validation as instructional formats continue to evolve. We therefore treat the framework as a living resource that should be updated periodically as new evidence accumulates, and we view the explicit tracking of context-dependence (Section 4.4.3) as a prerequisite for reliably distinguishing the stable core from the context-bound periphery in future revisions.

#### 4.4.6 Validation Status

The discriminative cues identified in Level 4 are derived analytically (by computing set differences) rather than empirically validated. They represent hypotheses about what cues should distinguish states, pending direct experimental confirmation.

### 4.5 Closing Remarks

The Nonverbal Syntax Framework addresses a fundamental challenge in affective computing and educational research: transforming scattered, inconsistently labeled findings into an integrated, usable resource. Through comprehensive normalization, dual-level evidence assessment (component and relationship tiers), actionability classification, and hierarchical organization, the framework enables researchers to identify knowledge gaps, practitioners to make evidence-calibrated inferences, and technologists to build more principled detection systems.

The framework identifies a critical insight often obscured in individual studies: while many cue-state relationships receive high confidence ratings based on well-studied components, only 4.5% of unique relationships (covering 35.5% of all mappings) have sufficient direct replication (R1–R4) to warrant confident practical application. By making this distinction explicit, the framework promotes appropriate epistemic humility while highlighting the 480 actionable relationships that practitioners can use with confidence.

The metaphor of *nonverbal syntax* captures the framework's core principle: just as understanding grammatical rules enables language comprehension, understanding the patterns by which behavioral cues combine to signal internal states enables more accurate state inference. Multi-cue patterns, the "sentences" of nonverbal communication, provide more reliable indicators than individual cues alone. The framework provides the "grammar book", a structured, evidence-graded catalog of these patterns, while acknowledging that individual expression, context, and continuing research will refine our understanding of this behavioral language.

## 5. CONCLUSION

This paper introduced the Nonverbal Syntax Framework, a systematic approach to organizing empirical knowledge about the behavioral manifestations of cognitive and affective states in learning contexts. Built on a comprehensive systematic literature review of 908 empirical studies yielding 17,043 cue-state mappings (Turaev et al., 2026), the framework addresses a fundamental challenge: transforming fragmented, inconsistently labeled research findings into an integrated, evidence-graded resource for both research and practice.

We developed systematic procedures that consolidated 5,537 raw state labels into 2,010 normalized states (63.7% reduction) and 11,521 raw cue descriptions into 6,434 normalized cues (44.2% reduction). The accompanying dictionaries enable consistent terminology across future research and reveal that apparent diversity in findings often reflects labeling differences rather than substantive disagreement.

The framework introduces two complementary evidence metrics. Component Evidence Tiers (Strong through Exploratory) assess how well-studied individual cues and states are across all their associations. Relationship Evidence Tiers (R1–R6) assess how well-documented each specific cue-state pair is. This dual assessment reveals a critical finding: 52% of relationships rated "Very High" confidence based on well-studied components are actually documented by only a single paper. By making this distinction explicit, the framework enables appropriately calibrated inference.

The framework distinguishes between generic cues requiring further specification (e.g., "facial expressions") and specific, observable behaviors that practitioners can directly detect and record (e.g., "furrowed brow," "scratching head," "sighing"). Over two-thirds of mappings (67.9%) involve cues that are at least moderately actionable, enabling their application in observation-based contexts. Cues are further classified by detection modality, distinguishing directly observable behaviors (across six channels: Facial, Head, Body, Gesture, Voice, and Eye in part) from instrumentally measured indicators (Physiology, Behavioral, Multimodal). This enables practitioners to filter the framework to cues appropriate for their detection context.

The hierarchical organization supports multiple use cases. Level 1 (Cue Vocabulary) answers "What might this behavior indicate?" Level 2 (State Clusters) answers "What cues indicate this state?" Level 3 (State Profiles) provides multimodal behavioral signatures with detailed, actionable cue specifications. Level 4 (Discriminative Analysis) resolves ambiguity between confusable states. The identification of 1,215 state pairs sharing common cues and the specification of discriminative cues for each pair address a critical gap. While prior work has documented cue-state associations, the framework uniquely explains how to differentiate between states with overlapping indicators.

A key outcome of this work is the identification of 480 actionable relationships with R1–R4 evidence (≥3 independent papers). While this represents only 4.5% of unique cue-state pairs, these relationships cover 35.5% of all documented mappings (6,049 of 17,043), span 47 states (including all key learning states such as engagement, confusion, frustration, boredom, and attention), include 111 distinct cues across observable behavioral indicators, and represent 8 of 9 behavioral channels. These 480 relationships constitute the consolidated, replicated core of six decades of research, examined across the 908 papers in the corpus. Practitioners can apply these relationships with confidence. Of the remaining unique relationships, 91.5% (9,653 single-paper findings) represent the exploratory frontier, that is, hypotheses for future replication that expand our understanding but should not yet guide confident practical application.

The framework embodies a core insight: internal states manifest through coordinated patterns of observable behavior, a *nonverbal syntax* that can be systematically decoded. Just as linguistic syntax enables meaningful communication through rule-governed word combinations, nonverbal syntax enables state inference through patterned behavioral combinations. The key finding supporting this principle is that multi-cue patterns provide more reliable state indicators than individual cues alone. A furrowed brow (AU4) appears in confusion (35 papers), frustration (15 papers), and engaged concentration (13 papers). Only by observing accompanying cues, such as scratching the head suggesting confusion, sighing suggesting frustration, and forward lean suggesting engagement, can reliable inference be achieved. Individual cues are the "words"; multi-cue patterns are the "sentences" of nonverbal communication.

Understanding this syntax involves recognizing which specific cues carry reliable diagnostic information (R1–R4 evidence), integrating evidence across behavioral channels through multimodal observation, applying discriminative reasoning when multiple states are consistent with observations, and calibrating confidence to relationship evidence strength rather than to component evidence alone.

For researchers, the framework identifies knowledge gaps (understudied states, channels, and relationships), provides standardized terminology, reveals that only 4.5% of documented relationships have sufficient replication (R1–R4) for confident application, and supports theory testing through structured evidence organization.

For practitioners (educators, trainers, clinicians), the framework offers evidence-calibrated guidance for observational assessment. The detailed behavioral signatures for key learning states specify exactly which observable behaviors to monitor, with explicit paper counts and evidence tiers. Verbal indicators ("I'm confused", "This is boring") provide unambiguous state identification when present.

For technologists developing affect-aware systems, the framework provides validated features for state detection (prioritizing R1–R3 relationship evidence), confidence calibration guidelines based on relationship tiers rather than component tiers alone, discriminative cues for resolving classification ambiguities, and clear guidance on which cues require instrumentation versus those detectable through standard cameras or microphones.

The framework synthesizes correlational evidence from heterogeneous studies and cannot establish causation. Individual differences, cultural factors, and contextual influences produce variation not captured in population-level patterns. The discriminative cues represent analytically derived hypotheses that require empirical validation. A detailed discussion of limitations appears in Section 4.4 and in Turaev et al. (2026).

Priority directions for future work include direct comparison studies testing discriminative cues between confusable states; multimodal sensing studies examining cross-channel pattern co-occurrence; validation of the relationship evidence tier system through predictive accuracy studies; extension to temporal dynamics and intensity gradations; cultural validation studies; and the development of practical tools for observation and automated detection.

The Nonverbal Syntax Framework represents a step toward making decades of affective computing and educational research more accessible and actionable. By organizing what we collectively know about how internal states manifest behaviorally, and critically, by distinguishing between well-replicated relationships and preliminary observations, the framework supports more principled research, more confident practice, and more informed technology development.

The framework reveals both the richness and the limitations of current knowledge. While 10,553 cue-state relationships have been documented, the 480 actionable relationships (4.5%) with R1–R4 evidence cover 35.5% of all mappings and span the key states, cues, and channels that practitioners need. This finding should guide both future research priorities (replicating promising relationships) and current practice (appropriately calibrating confidence across different evidence levels).

The framework and all supporting materials, including normalization dictionaries, tiered data structures, detailed behavioral signatures, and analysis code, are provided as open resources. We invite the research community to use, extend, and refine these tools as our collective understanding of the nonverbal syntax of learning continues to evolve.

## SUPPLEMENTARY MATERIALS

Supplementary materials, including data and code supporting this study, are available from the corresponding author upon reasonable request.